\documentclass[10pt,twocolumn,letterpaper]{article}

\usepackage{iccv}              

\definecolor{iccvblue}{rgb}{0.21,0.49,0.74}
\usepackage[pagebackref,breaklinks,colorlinks,allcolors=iccvblue]{hyperref}

\usepackage{amsmath}
\usepackage{booktabs}
\usepackage{multirow}
\usepackage{graphics,dblfloatfix}
\usepackage{url}
\usepackage{color, colortbl}
\usepackage{algorithm}
\usepackage{algpseudocode}
\usepackage{rotating}
\usepackage{caption}
\usepackage{subcaption}
\usepackage{pifont}
\usepackage{arydshln}

\usepackage{amsmath,amsfonts,bm}

\def\eqref#1{equation~\ref{#1}}

\def\1{\bm{1}}

\def\eps{{\epsilon}}

\DeclareMathAlphabet{\mathsfit}{\encodingdefault}{\sfdefault}{m}{sl}
\SetMathAlphabet{\mathsfit}{bold}{\encodingdefault}{\sfdefault}{bx}{n}

\DeclareMathOperator*{\argmax}{arg\,max}
\DeclareMathOperator*{\argmin}{arg\,min}

\let\originalleft\left
\let\originalright\right
\renewcommand{\left}{\mathopen{}\mathclose\bgroup\originalleft}
\renewcommand{\right}{\aftergroup\egroup\originalright}

\usepackage{amsmath}
\usepackage{amssymb}  

\usepackage{mathtools}  

\makeatletter
\newcommand{\spx}[1]{
	\if\relax\detokenize{#1}\relax
	\expandafter\@gobble
	\else
	\expandafter\@firstofone
	\fi
	{^{#1}}%
}
\makeatother

\usepackage{stmaryrd}  
\newcommand{\genericdel}[4]{%
	\ifcase#3\relax
	\ifx#1.\else#1\fi#4\ifx#2.\else#2\fi\or
	\bigl#1#4\bigr#2\or
	\Bigl#1#4\Bigr#2\or
	\biggl#1#4\biggr#2\or
	\Biggl#1#4\Biggr#2\else
	\left#1#4\right#2\fi
}

\usepackage{bm}
\usepackage[OMLmathsfit]{isomath}  
\usepackage{upgreek}

\DeclareMathAlphabet{\mathbbmrm}{U}{bbm}{m}{rm}
\DeclareMathAlphabet{\mathbbmsl}{U}{bbm}{m}{sl}
\DeclareMathAlphabet{\mathbbmb}{U}{bbm}{b}{it}
\DeclareMathAlphabet{\mathbbmssit}{U}{bbmss}{m}{it}

\let\vec\relax

\newcommand{\vec}[1]{\bm{#1}}

\newcommand{\nunder}[2][5]{\mathrlap{\mkern\the\numexpr#1/2mu\relax\underline{\phantom{\mathrm{#2}\mkern-#1mu}}}\mathrm{#2}}

\let\oldhat\hat
\renewcommand{\hat}[1]{\vphantom{#1}\smash[t]{\oldhat{#1}}}
\let\oldtilde\tilde
\renewcommand{\tilde}[1]{\vphantom{#1}\smash[t]{\oldtilde{#1}}}
\let\oldwidetilde\widetilde
\renewcommand{\widetilde}[1]{\vphantom{#1}\smash[t]{\oldwidetilde{#1}}}

\newenvironment{talign*}
{\csname align*\endcsname}
{\endalign}

\usepackage{dashbox}%

\definecolor{Gray}{gray}{0.9}

\def\paperID{12702}
\def\confName{ICCV}
\def\confYear{2025}

\title{Seal Your Backdoor with Variational Defense}

\author{Ivan Sabolić \quad \quad Matej Grcić \quad \quad Siniša Šegvić \\
University of Zagreb, Faculty of Electrical Engineering and Computing\\
Unska 3, 10000 Zagreb, Croatia\\
{\tt\small name.surname@fer.hr}
}

\begin{document}
\maketitle


\begin{abstract}
We propose VIBE, 
a model-agnostic framework 
that trains classifiers 
resilient to backdoor attacks.
The key concept behind our approach
is to treat malicious inputs 
and corrupted labels 
from the training dataset 
as observed random variables,
while the actual clean labels
are latent.
VIBE then recovers the corresponding latent clean label posterior
through variational inference. 
The resulting training procedure 
follows the expectation-maximization (EM) algorithm.
The E-step infers the clean pseudolabels by solving
an entropy-regularized optimal transport problem,
while the M-step updates the classifier parameters via gradient descent.
Being modular,
VIBE can seamlessly integrate
with recent advancements 
in self-supervised representation learning,
which enhance its ability 
to resist backdoor attacks.
We experimentally validate the method effectiveness 
against contemporary backdoor attacks 
on standard datasets, a large-scale setup with 1$k$ classes,
and a dataset poisoned with multiple attacks.
VIBE consistently outperforms 
previous defenses 
across all tested scenarios. 
\end{abstract}

\section{Introduction}


Deep models possess enough capacity to learn any pattern present within the data \cite{zhang17iclr,kolesnikov20eccv,brown20neurips}.
This remarkable flexibility comes at the cost of control since it limits our ability to influence 
the specific motifs the model learns \cite{arpit17icml,kalimeris19neurips}.
For instance, a model may base its decisions on image backgrounds rather than focusing on the actual objects \cite{geirhos20nmi}.
Such bias towards simpler \cite{shah20neurips} and possibly spurious patterns \cite{izmailov22neurips} may lead to undesirable generalizations that reveal themselves only in specific test cases.
This deep learning loophole can be maliciously exploited by attackers who manipulate training examples using triggers that steer the model towards harmful generalization.
Such practice is commonly referred to as \textit{backdoor learning} \cite{li2022ieee} and presents a serious threat in contemporary machine learning.


The majority of existing backdoor attacks \cite{gu2019ieee,chen2017arxiv} 
modify a portion of the training dataset by installing triggers onto selected inputs and 
altering the corresponding labels\footnote{Some attacks do not alter the labels \cite{turner2019arxiv}. However, our experiments show that they are much easier to defend from.}.
On such data, standard supervised learning delivers a \textit{poisoned} model \cite{li2022ieee}.
During inference, attackers can exploit the installed backdoor by applying triggers to the desired inputs, which causes the model to behave maliciously \cite{gu2019ieee,nguyen2021iclr}.
Our goal is to defend against such attacks by training a \textit{clean} model 
invariant to triggers present in the data.

Recent empirical defenses \cite{chen2022neurips,liu2023iccv,zhu23iccv} partition the training dataset into clean and poisoned subsets according to some heuristics.
The two subsets then take different roles during the model training (\textit{e.g.}~semi-supervised learning with labeled clean data and unlabeled poisoned data \cite{huang2022iclr}).
However, heuristics are prone to failure modes and can be exploited by adaptive attacks \cite{qi2022iclr}.
Also, pruning labels often leads to information loss, ultimately degrading recognition performance.
Our approach avoids data partitioning and label pruning.
Instead, we leverage optimal transport
to refine potentially corrupted samples and labels
into clean pseudolabels that guide the training of a robust classifier.


In this work, we present VIBE (\textbf{V}ariational \textbf{I}nference for \textbf{B}ackdoor \textbf{E}limination), a framework for training backdoor-robust classifiers on poisoned data.
Our key concept is to treat dataset examples and the corresponding corrupted labels as observed random variables, while the desired clean labels are latent.
Then, we achieve resilience against backdoor attacks by recovering the latent clean posterior parametrized as a deep classifier.
VIBE training takes the form of an 
expectation-maximization
algorithm that alternates between classifier updates via gradient descent (M-step), and inference of approximate clean class posterior (E-step).
In practice, the approximate clean labels are recovered by solving an entropy-regularized optimal transport problem \cite{cuturi13neurips}.
We validate the resilience against contemporary backdoor attacks on standard benchmarks, on a large-scale setup with 1k classes, and on a dataset poisoned with multiple attacks.
Experiments indicate consistent improvements over previous defenses in all tested scenarios.
Remarkably, VIBE attains over 12pp absolute improvement in ASR over the best baseline on the CIFAR-10 dataset. 

\section{Related work}

\textbf{Backdoor attacks.} 
Backdoor attacks achieve malicious model behaviour through direct
modifications of trainable parameters~\cite{rakin2020cvpr}, changes in model structure~\cite{hong2022neurips}, or data poisoning~\cite{gu2019ieee}.
Contemporary machine learning models are trained in-house and deployed via APIs which makes parameter- and structure-based attacks largely impractical.
Therefore, we focus on a more realistic scenario where the attacker only controls the data collection process.

Early data poisoning attacks \cite{gu2019ieee} steer the model towards malicious generalization by introducing localized triggers and altering the corresponding labels.
Subsequent attacks rely on  invisible~\cite{cheng2021aaai,
qi2022iclr, wang2022eccv, jiang2023cvpr} or sample-specific~\cite{li2021cvpr, zhang2022ieee} triggers, which are significantly harder to detect.
Clean-label attacks~\cite{turner2019arxiv, barni2019ieee, yu2024icml} avoid modifying labels altogether but typically deliver lower attack success rates. 
All these approaches devise broad method-agnostic attacks.
Contrary, recent adaptive attacks improve effectiveness by targeting the latest defenses \cite{huang2022iclr,gao2023cvpr}.

Another line of work considers data poisoning attacks that target contrastive self-supervised learning \cite{saha22cvpr,li23iccv,liang24cvpr}.
However, these attacks may require access to the  optimization procedure \cite{jia22sp} and typically have a lower success rate than direct attacks on supervised learning \cite{li23iccv}.
A detailed survey of backdoor attacks can be found in~\cite{li2022ieee}.


\noindent
\textbf{Backdoor defenses.}
Existing defenses can be categorized as either certified or empirical. 
Certified defenses provide 
theoretical guarantees of success \cite{xiang2023neurips}.
However, their underlying assumptions
typically do not hold in practice \cite{li2022ieee}.
Empirical defenses devise preprocessing strategies to avoid training on corrupted examples \cite{tran2018neurips}, correct the malicious generalization of poisoned models via postprocessing \cite{liu2017iccd}, or propose heuristic additions to standard training algorithms \cite{li2021neurips}.
Preprocessing-based defenses~\cite{chen2019aaaiw, guo2023iclr,huang2023iclr,khaddaj2023icml}
aim to filter out poisoned examples from the dataset, allowing the remaining clean data to be safely used for supervised training.
Such methods cannot distinguish
poisoned examples from their
hard counterparts with the correct label \cite{khaddaj2023icml}.

Post-training defenses \cite{zeng2022iclr, li2021iclr,Pang2022cvpr,xie2024iclr} focus on removing backdoors from already trained models.
A common approach involves re-synthesizing the injected triggers and using them to purify the model~\cite{wang2019sp, qiao2019neurips, zhu2020acm, shen2021icml, guo2022iclr, xiang2022iclr, Liu2022cvpr, Tao2022cvpr, wang2023iclr, xu2024iclr}.
Other approaches correct malicious generalization through 
model pruning~\cite{liu2018israid, wu2021neurips, Li2023icml}, knowledge distillation \cite{yoshida2020acm}, loss landscape analysis \cite{zhao2020iclr}, or by enhancing robustness against adversarial examples \cite{Mu2022cvpr}.
All these methods assume access to a small subset of definitively clean data, which may not be available in practice.

 Training-time defenses~\cite{huang2022iclr, chen2022neurips, gao2023cvpr, zhang2023cvpr, liu2023iccv,zhu23iccv, zhao25aaai, wei2024neurips, Sabolic2024BMVC} attempt to train robust classifiers from poisoned data.
An early approach \cite{li2021iclr} isolates poisoned examples in early training stages and later uses them to unlearn the backdoors.
Subsequent methods~\cite{huang2022iclr, gao2023cvpr, chen2022neurips} 
focus on removing only the labels of potentially poisoned samples and proceed with semi-supervised training.
These approaches identify the poisoned data by heuristics, which increases the defense vulnerability.
VIBE avoids such heuristics by recovering clean pseudolabels through variational inference.

\noindent
\textbf{Representation learning.}
The main goal of representation learning \cite{rumelhart86n,wang15iccv,pathak16cvpr} is to recover features that generalize across a spectrum of downstream tasks.
A widely used representation learning strategy involves optimizing self-supervised pretext objectives \cite{noroozi16eccv,gidaris18iclr}.
Recent such methods
reconstruct masked inputs \cite{he22cvpr}, optimize contrastive objectives \cite{chen2020simple} or learn latent centroids \cite{oquab24tml}.
Representation quality can be further enhanced by training on large multimodal datasets \cite{radford21icml},
which delivers effective features even without fine-tuning on target datasets \cite{cherti23cvpr,oquab24tml}.

Representation learning received limited attention in the context of backdoor defenses.
Initial works design heuristics that leverage self-supervised representations to remove poisoned labels \cite{huang2022iclr} or filter the dataset \cite{wang2023icassp}.
These heuristics fail for some attacks, as indicated by our experimental evaluation.
In contrast, VIBE uses self-supervised pre-training to jumpstart the optimization process. 

Recent works \cite{yang24icml, xun2024arxiv, bie2024mitigating, han2024mutual} suggest that (multimodal) contrastive learning can be conveniently adapted to resist data poisoning attacks that target self-supervised pre-training.
Pre-training VIBE feature extractor according to such objectives further boosts the performance.

\noindent
\textbf{Latent variable models.}
Latent variable models \cite{everitt84book} explain relations between observed random variables with latent variables.
This concept is successfully applied in different fields \cite{kingma14iclr,lopez18nm,wright20chrmcp}.
VIBE introduces latent variables into backdoor defenses by viewing clean labels as latents.

\section{Backdoor resilience via variational inference}


\textbf{Problem setup.}
Let $\mathcal{D}_\text{raw} = \left\{( \tilde{\vec{x}}^i, l^i )\right\}_{i=1}^{N}$
be a benign dataset consisting of input examples $\tilde{\vec{x}}^i \in \mathcal{X}$ and clean labels $l^i \in \mathcal{Y}$, where $\mathcal{X}$ and $\mathcal{Y}$ are input and label space respectively. 
A malicious attacker $\tau: \mathcal{X} \times \mathcal{Y} \rightarrow \mathcal{X} \times \mathcal{Y}$
with a budget $\gamma \in [0, 100]$ modifies $ \gamma\%$ examples by triggering inputs and corrupting their labels.
The remaining $(100-\gamma)\%$ of the data remain unchanged and correctly labeled in order to conceal the attack.
Given a corrupted dataset $\mathcal{D} = \tau(\mathcal{D}_\text{raw}) = \{(\vec{x}^i, y^i)\}_{i=1}^N$, our goal is to train a robust classifier $f: \mathcal{X} \rightarrow \mathcal{Y}$ that assigns clean labels $l^i \in \mathcal{Y}$ to every input $\vec{x}^i$ while ignoring the malicious triggers.

The core concept behind VIBE is to treat clean labels as unobserved latent variables $\underline{l}$.
We then frame the training of a clean classifier as a latent posterior recovery from observed inputs $\underline{\vec{x}}$ and corrupted labels $\underline{y}$.

Backdoor attacks typically poison a small portion of the data in order to stay undetected~\cite{li2022ieee}. In our framework, this means that corrupted and clean labels are often identical.
Thus, a natural optimization objective is to maximize the conditional log-likelihood of the i.i.d dataset $\mathcal{D}$ given a set of model parameters $\theta$:
\begin{equation}
\label{eq:vd_obj}
\ell(\theta|\mathcal{D}) 
= \ln \prod_{i=1}^N p_\theta(y^i| \vec{x}^i) 
= \sum_{i=1}^N \ln \sum_{l=1}^K p_\theta(y^i| l, \vec{x}^i) \,  p_\theta(l| \vec{x}^i).
\end{equation}
For simplicity, we abbreviate $p(\underline{l} = l| \underline{\vec{x}} = \vec{x}^i)$ as $p(l|\vec{x}^i)$.
Given the likelihood factorization (\ref{eq:vd_obj}), we proceed by deriving a tractable optimization objective.
Note that we defer concrete parametrization of the clean class posterior $p_\theta(\underline{l}| \underline{\vec{x}})$ and the corrupted class posterior $p_\theta(\underline{y}|\underline{l},\underline{\vec{x}})$ to Section~\ref{subsec:paramet_post}.

\subsection{Optimizing the variational objective via EM}
Direct maximization of $\ell(\theta|\mathcal{D})$
does not ensure the correct recovery
of the clean class posterior since the clean class is latent \cite{mcLachlan08book}.
Fortunately, we can turn to variational inference and maximize likelihood lower bound $\ell_\text{ELBO}$ that 
introduces an approximate latent posterior $q$:
\begin{align}
    \ell(\theta|\mathcal{D}) 
    &= \sum_{i=1}^N \ln \sum_{l=1}^K p_{\theta}(y^i | l, \vec{x}^i) \, p_{\theta}(l | \vec{x}^i) \frac{q(l|\vec{x}^i, y^i)}{q(l|\vec{x}^i, y^i)} \nonumber\\
    &\geq \sum_{i=1}^N \mathbb{E}_{l^i \sim q(\cdot|\vec{x}^i, y^i)} \left[ \ln \frac{p_{\theta}(y^i | l^i, \vec{x}^i) p_{\theta}(l^i | \vec{x}^i)}{ q(l^i|\vec{x}^i, y^i)} \right] \nonumber \\ &=: \ell_\text{ELBO}(\theta, q|\mathcal{D}).
\label{eq:vd_elbo}
\end{align}
The inequality follows directly from Jensen's inequality.
We optimize the proposed $\ell_\text{ELBO}$ objective with the expectation-maximization (EM)
algorithm \cite{mcLachlan08book}.
In practice, this involves alternating between 
updates of the approximate latent posterior $q$ (E-step) and parameters $\theta$ (M-step).

\noindent
\textbf{E-step:~updating the approximate latent posterior.}
We begin by observing that the $\ell_\text{ELBO}$ objective 
requires only the recovery of $q$ for dataset examples, rather than an exact closed-form distribution.
With this observation in mind, we rewrite the objective (\ref{eq:vd_elbo}) averaged over $N$ examples as:
\begin{align}
   \frac{1}{N} \ell_\text{ELBO} &=  \sum_{i=1}^N \sum_{l=1}^K \frac{1}{N} q(l|\vec{x}^i, y^i) \ln [p_{\theta}(y^i | l, \vec{x}^i) p_{\theta}(l | \vec{x}^i)] \nonumber \\ &- \sum_{i=1}^N \sum_{l=1}^K \frac{1}{N} q(l|\vec{x}^i, y^i)\ln  q(l|\vec{x}^i, y^i).
\end{align}
We next substitute  $\mathbf{P}_{i, l} := p_{\theta}(y^i | l, \vec{x}^i) p_{\theta}(l | \vec{x}^i)$ and $\mathbf{Q}_{i, l} :=  \frac{1}{N} q(l|\vec{x}^i, y^i)$, where $1/N$
ensures that $\mathbf{Q}$ is a proper joint distribution \cite{asano20iclr,ni22iclr}. 
Replacing the summations with matrix multiplication reveals the   same objective in the matrix form:
\begin{align}
   \frac{1}{N} \ell_\text{ELBO} &= \text{tr}(\textbf{Q}^\top \ln \textbf{P}) + \mathbb{H}(\textbf{Q}) +1 - \ln N \nonumber \\ &\geq \text{tr}(\textbf{Q}^\top \ln \textbf{P}) + \frac{1}{\lambda} \mathbb{H}(\textbf{Q}) + 1 - \ln N \, \label{eq:ot}
\end{align}
Here, $\text{tr}(\cdot)$ is the matrix trace operator,   $\lambda > 1$ is a hyper-parameter, and 
$\mathbb{H}(\mathbf{Q})$
is the entropy of coupling matrix $\mathbf{Q}$ \cite{peyre19ftml}. 
The complete derivation is deferred to Appendix~\ref{app:derivation_estep}.
The term $1-\ln N$ is constant and 
thus can be ignored.

Each matrix row $\mathbf{Q}_{i,:}$ sums to $1/N$ by the definition of $\mathbf{Q}$, while columns sum to the prior over clean classes $\boldsymbol{\pi}$.
Consequently, the set of all possible solutions for the objective (\ref{eq:ot}) forms a polytope:
\begin{equation}
\label{eq:polytope}
    \mathcal{Q}[\boldsymbol{\pi}] = \{ \, \textbf{Q} \in \mathbb{R}^{N \times K}_+ \, | \, \textbf{Q}^\top \mathbf{1}_N = \boldsymbol{\pi}, \,\,  \textbf{Q} \mathbf{1}_K = \frac{1}{N}  \mathbf{1}_N \, \}.
\end{equation}
Here, $\mathbf{1}_N$ is an $N$-dimensional column vector.
Maximizing the objective (\ref{eq:ot}) over $\mathcal{Q}[\boldsymbol{\pi}]$ is equivalent to solving the entropy-regularized optimal transport problem \cite{cuturi13neurips,peyre19ftml,asano20iclr}:
\begin{equation}
\label{eq:E_step}
   \mathbf{Q}^* = \argmin_{\textbf{Q} \in \mathcal{Q}[\boldsymbol{\pi}]} \left( \text{tr}(\textbf{Q}^\top \textbf{M}) - \frac{1}{\lambda} \mathbb{H}(\textbf{Q})\right).
\end{equation}
Here, the cost matrix contains the model outputs in dataset examples ($ \mathbf{M} = -\ln \textbf{P}$).
The optimal solution $\mathbf{Q}^*$ can be efficiently 
obtained with the Sinkhorn-Knopp's matrix scaling algorithm \cite{knopp67jm,peyre19ftml}, which we revisit in 
Appendix~\ref{app:sinkhorn}.
This approach is computationally efficient even for large $N$, as discussed in the experiments.
The recovered solution $\mathbf{Q}^*$ contains outputs of the approximate posterior $q$ for the dataset examples and allows us to proceed with the M-step.

\noindent
\textbf{M-step: updating model parameters.}
Given the outputs of approximate posterior $q$, we can turn to the optimization of parameters $\theta$.
Maximizing the  $\ell_\text{ELBO}$ objective (\ref{eq:vd_elbo}) is equivalent to the following minimization problem:
\begin{equation}
\label{eq:m_step}
    \min_{\theta}  \sum_{i=1}^N \text{CE}[q \, || \, p_{\theta}(l^i | \vec{x}^i)] + \mathbb{E}_{l^i \sim q} \left[- \ln p_{\theta}(y^i | l^i, \vec{x}^i) \right] 
\end{equation}
Here, CE denotes the cross-entropy loss. 
The full derivation can be found in Appendix~\ref{app:derivation_mstep}.
The objective (\ref{eq:m_step}) is continuous w.r.t parameters $\theta$ and can be optimized by (stochastic) gradient descent.
The rewritten objective highlights the role of the approximate posterior $q$: it acts as a pseudolabel generator.
These pseudolabels are also conditioned on the corrupted labels and thus provide a learning signal for the actual clean posterior.
The second objective term models the relation 
between the corrupted labels and the pseudolabels.
This term presents an opportunity 
to uncover the attacker's poisoning patterns
that can guide human inspection.


\begin{figure*}[htb]
    \centering
    \includegraphics[width=\linewidth]{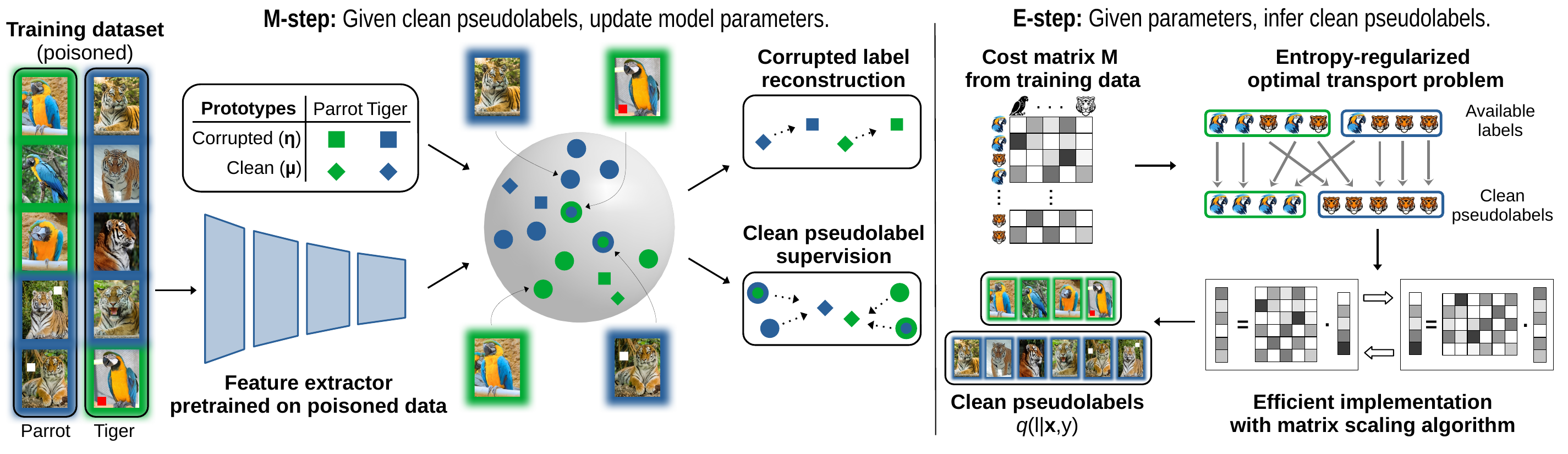}
    \caption{VIBE training alternates between iterative parameter updates (M-step) and inference of clean pseudolabels by solving entropy-regularized optimal transport problem efficiently implemented with matrix scaling algorithm (E-step).
    }
    \label{fig:train}
\end{figure*}

Altogether, VIBE training alternates between the described E and M steps as visualized in Figure~\ref{fig:train}.
The full algorithm is in Appendix~\ref{app:train_alg}.
Next, we discuss the implementations of distributions parameterized with $\theta$.

\subsection{Parameterizing the posteriors}
\label{subsec:paramet_post}
Let $g_{\theta_\text{E}}: \mathcal{X} \rightarrow S^{d-1}$ be a feature extractor that encodes inputs onto a ($d-1$)-dimensional unit hypersphere, \textit{e.g.}~a ResNet with $L_2$-normalized outputs.
We model the conditional likelihood of the encoded input $\vec{v}^i = g_{\theta_\text{E}}(\vec{x}^i)$ given the clean class $l^i$ with a von Mises-Fisher distribution~\cite{banerjee05jmlr}:
\begin{equation}
    p_{\theta}(\vec{v}^i | l^i)  = C_d(\kappa) \exp(\kappa \boldsymbol{\mu}_{l^i}^\top \vec{v}^i ).
\end{equation}
The vector $\boldsymbol{\mu}_{l^i} \in S^{d-1}$ sets the mean direction, the hyper-parameter $\kappa$ controls the distribution spread, while $C_d(\kappa)$ is a normalization constant \cite{banerjee05jmlr}. 
We can derive the clean label posterior as a vMF mixture via the Bayes rule:
\begin{equation}
\label{eq:true_posterior}
    p_{\theta}(l^i | \vec{x}^i) = 
    \frac{\exp( \kappa\,{\vec{v}^i}^\top \boldsymbol{\mu}_{l^i} + \ln \pi_{l^i})}{ \sum_{l'} \exp( \kappa\,{\vec{v}^i}^\top \boldsymbol{\mu}_{l'} + \ln \pi_{l'})}.
\end{equation}
Here, the mixing coefficient $\boldsymbol{\pi}$ induces a prior over clean classes.
In practice, we compute $\boldsymbol{\pi} = \sigma(c \cdot \theta_{\pi})$,
where $\sigma$ is softmax activation that ensures $\boldsymbol{\pi}$ is a distribution, $c$ is a hyperparmeter,
and  $\theta_{\pi} \in \mathbb{R}^d$ are learnable parameters.
The full derivation 
is in Appendix \ref{app:clean_posteriror}.
The clean posterior (\ref{eq:true_posterior}) corresponds to a softmax-activated deep model
with $L_2$-normalized pre-logits and clean class prototypes $\theta_{l} = \{\boldsymbol{\mu}_1, \dots, \boldsymbol{\mu}_K \}$.
Thus, 
the clean labels can be recovered by $f = \text{argmax} \circ \text{cos-sim}_{\theta_{l, \pi}} \circ g_{\theta_\text{E}}$, where $\text{cos-sim}$ operator computes 
cosine similarities
adjusted by the bias $\ln\pi$.

We model the corrupted class posterior $p_\theta(y^i|l^i, \vec{x}^i)$ as  cosine similarity between the corrupted class prototypes $\theta_y = \{\boldsymbol{\eta}_1, \dots, \boldsymbol{\eta}_K \}$ and output of function $h$ that process the encoded input $\vec{v}^i$ and the clean label prototype $\boldsymbol\mu_{l^i}$: 
\begin{equation}
\label{eq:pois_post}
    p_{\theta}(y^i | l^i, \vec{x}^i) := \frac{\exp(\nu \cdot {\boldsymbol{\eta}_{y^i}}^\top h(\boldsymbol{\mu}_{l^i}, \vec{v}^i))}{ \sum_{y'} \exp(\nu \cdot {\boldsymbol{\eta}_{y'}}^\top h(\boldsymbol{\mu}_{l^i}, \vec{v}^i))}.
\end{equation}
Here, $\nu$ is a scalar hyper-parameter, while details on $h$ are deferred to implementation details.
Note that the full corrupted posterior can be approximated as $p_{\theta}(y^i | l^i, \vec{x}^i) \approx p_{\theta}(y^i | l^i)$ by replacing output of $h$ with $\boldsymbol\mu_{l^i}$.
The detailed description of the approximated corrupted posterior is in Appendix~\ref{app:corrupted_posteriror}.
While this approximation makes optimization more challenging,
it enables seamless reconstruction of the systematic poisoning rules of the attacker $\tau$.
We experimentally evaluate both the full and approximate posterior.

Alltogether, the set of free parameters is a union $\theta = \theta_\text{E} \cup \theta_l \cup \theta_\pi \cup \theta_y$.
Next we analyze convergence of the EM algorithm with the introduced parametrization.

\subsection{Steering the EM algorithm convergence}
Our E-step solves a convex optimization problem \cite{cuturi13neurips}, while the M-step conducts non-convex training of a deep model. 
As a result, the EM algorithm may end up in a suboptimal stationary point
\cite{wu83as,mcLachlan08book,balakrishnan17as}.
In fact, the convergence point of the EM algorithm strongly depends on the initialization \cite{michael16adac}.
Fortunately, recent works observe that self-supervised pre-training of feature extractors ~\cite{chen2020simple,he20cvpr,estepa2023iccv}  lowers the sample complexity of the downstream task \cite{lee21neurips,alon24iclr} and improves generalization \cite{wang24iclr}.

\begin{figure}[htb]
    \centering    \includegraphics[width=0.95\linewidth]{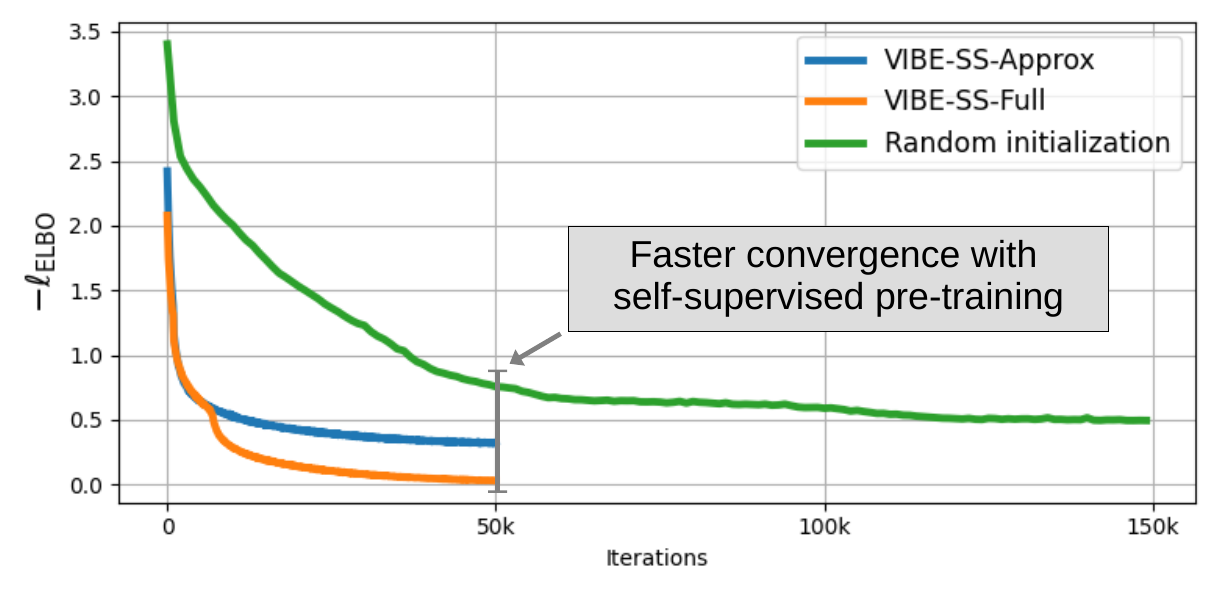}
  \caption{
  VIBE achieves faster convergence and improved generalization with self-supervised pre-training.
  }
    \label{fig:losses_and accuracies}
\end{figure}

Therefore, we conduct self-supervised pre-training on the poisoned dataset instance (similar to \cite{huang2022iclr}) before end-to-end optimization of our $\ell_\text{ELBO}$ objective.
Figure~\ref{fig:losses_and accuracies} shows that self-supervised pre-training on poisoned data
jumpstarts the EM optimization, leads to faster convergence, and increases the likelihood lower bound $\ell_{\text{ELBO}}$. 
Moreover, the corresponding solution generalizes better and turns out to be near optimal compared to supervised learning on clean labels.
Still, the self-supervised pre-training does not compromise the generality of VIBE, as pre-training objectives are already available for various modalities \cite{yoon20neurips,kawakami20emnlp,han22bb}.

Scaling the concept of feature extractor pre-training in terms of dataset size leads to foundation models \cite{radford21icml,cherti23cvpr,oquab24tml} like CLIP and DINOv2.
Modular design of VIBE posteriors enables integration of these off-the-shelf extractors, allowing performance analysis in the \textit{transfer learning} setup.
Nevertheless, foundation models should be carefully downloaded from trusted third-party providers or pre-trained with robust procedures \cite{yang24icml, xun2024arxiv} that avoid backdoor injection during self-supervised pre-training stage.

\subsection{Clean-label attacks diverge from data manifold}
\label{sec:clean_label_manifold}

\begin{figure}[h]
    \centering
    \includegraphics[width=\linewidth]{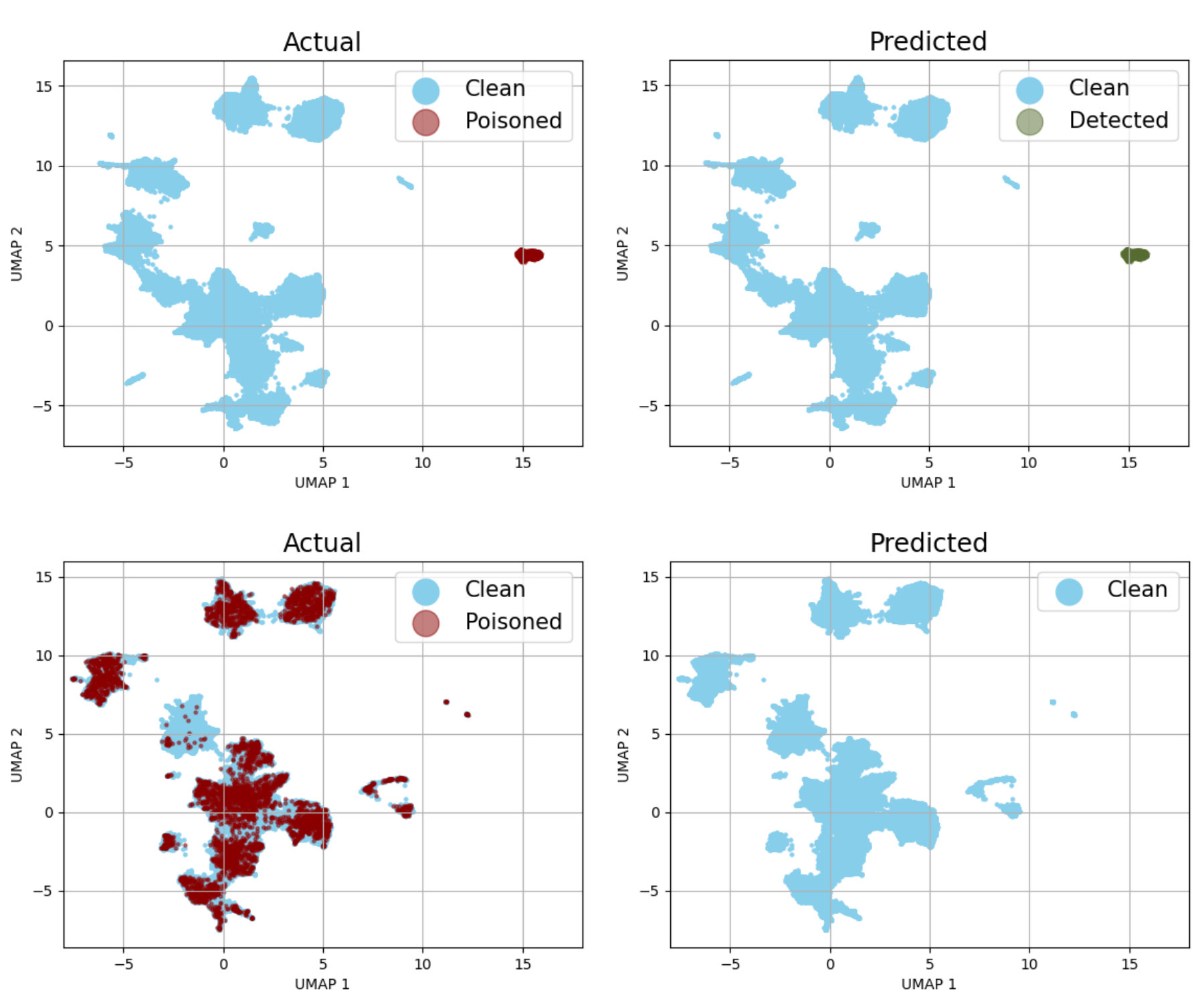}
    \caption{
    Clean-label attacks shift the poisoned examples away from the clean data manifold (top-left). Thus, we can detect them as outlier communities (top-right). 
    Poisoned-label attacks remain on the manifold and can be useful for model training (down-left).
    Thus,
    our preprocessing strategy keeps them (down-right).
    } 
    \label{fig:feature_space}
\end{figure}

Poisoned-label attacks strive for stealthiness by injecting minimal triggers.
Thus, the poisoned examples remain near the clean data manifold.
However, 
clean-label attacks \cite{turner2019arxiv, barni2019ieee} operate by significantly perturbing the inputs to construct a successful attack.
These perturbations shift the poisoned examples away from the data manifold in the self-supervised feature space, as illustrated in Figure \ref{fig:feature_space}.
We propose a pre-processing technique 
that
exploits this property of clean-label attacks
to identify the poisoned examples.

Our preprocessing strategy captures the data manifold within a self-supervised feature space and identifies perturbed examples as outlier communities.
Specifically, we construct $k$ nearest neighbors graph, represented by the adjacency matrix $A_k$.
Then, we apply a community detection algorithm \cite{traag19sr} that partitions graph $A_k$ into $K$+1 distinct communities.
We proceed by computing pairwise distances between the communities to identify the most distant one.
Finally, we remove the most distant community if its average distance to other communities exceeds some predefined threshold $\delta$.
In the case of clean-label attacks, the most distant community corresponds to off-manifold data which is then removed. 
If poisoned examples are absent from the data, the most distant community is retained due to being much closer to the data manifold.
While some clean examples may be lost if $\delta$ is poorly selected, they typically do not affect the model generalization.
More details on preprocessing strategy are in Appendix~\ref{app:manifold_preprocess}.


\section{Experimental setup}

\textbf{Datasets \& metrics.}
We evaluate VIBE on the standard backdoor benchmarks: 
CIFAR-10, CIFAR-100 \cite{krizhevsky12}, and a subset of 30 classes from the ImageNet-1k dataset \cite{deng09cvpr}.
Furthermore, we scale the problem by considering the full ImageNet dataset with $1$k classes and $1.2$M training examples.
We evaluate the performance with the standard metrics: accuracy on the clean test set labels (ACC), and the attack success rate (ASR).
Lower ASR indicates more accurate recognition of poisoned samples and better defense.

\noindent
\textbf{Attacks.}
We consider eight attacks that represent the major backdoor attack families.
For visible patch-like attacks, we include BadNets~\cite{gu2019ieee} and Adap-Patch~\cite{qi2022iclr}).
For invisible attacks, we consider Blend~\cite{chen2017arxiv}, WaNet~\cite{nguyen2021iclr}, Adap-Blend~\cite{qi2022iclr} and Frequency \cite{wang2022eccv}. 
Clean-label attacks are represented by LC~\cite{turner2019arxiv}.
We also validate the \textit{all-to-all}
attack~\cite{li2022ieee} using a variant of BadNets.
By default, the poisoning rate is set to $10\%$ 
except for Adap-Patch and Adap-Blend attacks in which $1\%$ of the data is poisoned as suggested in \cite{qi2022iclr}.
Also, in the case of clean-label attacks, we poison $2.5\%$ of the data as suggested in \cite{huang2022iclr,gao2023cvpr}.
We set the target label as the zeroth class except for the \textit{all-to-all} attack.
As observed in previous works \cite{li2021neurips, chen2022neurips}, some attacks cannot be reproduced for all datasets.
Thus, experiments conducted on ImageNet-30 and CIFAR-100 show a subset of attacks.
Detailed configurations are in Appendix~\ref{app:attacks_details}.

\begin{table*}[b]
\setlength{\tabcolsep}{3.5pt}
    \centering
    \footnotesize
    \begin{tabular}{cccccccccccccc|cccc}
    \hline
         \multirow{2}{*}{Data} & Defense $\rightarrow$ &
         \multicolumn{2}{c}{No Defense} &
         \multicolumn{2}{c}{ABL} &
         \multicolumn{2}{c}{DBD} &
         \multicolumn{2}{c}{CBD} &
         \multicolumn{2}{c}{ASD} &
         \multicolumn{2}{c|}{VAB} &
         \multicolumn{2}{c}{VIBE-SS-A} &
         \multicolumn{2}{c}{VIBE-SS-F} \\ 
         & Attack $\downarrow$ & ACC & ASR & ACC & ASR & ACC & ASR & ACC & ASR & ACC & ASR & ACC & ASR & ACC & ASR & ACC & ASR \\ \hline
        \multirow{10}{*}{\begin{turn}{90}CIFAR-10\end{turn}} & No Attack & 95.0 & -  & 85.2 & - & 91.6 & - & 91.3 & - & 93.3  & - & 94.5 & - & 94.4 & - & 94.7 & - \\ \cdashline{2-16}[0.5pt/5pt]
        
        ~ & BadNets & 94.9 & 100 & 93.8 & 1.1 & 92.4 & 1.0 & 91.8 & 1.2 & 92.1 & 3.0 & 93.5 & 0.7 &  94.4 & 0.6 & 94.4 & 0.1 \\ 
        ~ & Blend & 94.2 & 98.3 & 91.9 & 1.6 & 92.2 & 1.7 & 90.0 & 96.6 & 93.4 & 1.0 & 93.9 & 0.4 & 93.6 & 8.7 & 94.6 & 0.0 \\ 
        ~ & WaNet & 94.3 & 98.0 & 84.1 & 2.2 & 91.2 & 0.4 & 91.6 & 97.3 & 93.3 & 1.2 & 94.2 & 0.5 & 94.1 & 0.9 & 94.3 & 0.7 \\ 
        ~ & Frequency & 94.9 & 100 & 81.3 & 8.8 & 92.3 & 2.6 & 91.6 & 100 & 88.8 & 100 & 93.8 & 0.4 &  94.1 & 0.8 & 94.4 & 0.0 \\ 
        ~ & Adap-Patch & 95.2 & 80.9 & 81.9 & 0.0 & 92.9 & 1.8 & 91.6 & 97.8 & 93.6 & 100 & 94.3 & 1.1 & 94.3 & 1.1 & 94.5 & 8.6 \\ 
        ~ & Adap-Blend & 95.0 & 64.9 & 91.5 & 81.9 & 90.1 & 99.9 & 92.3 & 87.5 & 94.0 & 93.9 & 94.5 & 29.1 & 94.5 & 36.7 & 94.5 & 9.0 \\ 
        ~ & LC & 94.9 & 99.9 & 86.6 & 1.3 & 89.7 & 0.0 & 91.3 & 24.7 & 93.1 & 0.9 & 94.0 & 16.6 & 93.2 & 5.3 & 93.0 & 6.0 \\ 
        ~ & BN-all2all & 92.2 & 91.5 & 91.2 & 0.4 & 92.9 & 0.6 & 92.6 & 91.9 & 93.6 & 2.2 & 94.5 & 92.2 & 94.3 & 0.6 & 94.6 & 1.2\\[0.2em]
        \rowcolor{Gray}
        \cellcolor{white} &  Average & 94.5 & 90.5 & 87.3 & 16.0 & 91.4 & 17.7 & 91.4 & 84.0 & 92.8 & 42.7 & 94.1 & 18.0  &\textbf{94.1} & \textbf{6.8} & \textbf{94.3} & \textbf{3.2} \\ 
        \hline

        \multirow{6}{*}{\begin{turn}{90}CIFAR-100\end{turn}} & No Attack & 74.9 &  -& 70.5 &-  & 66.2  &- & 71.1  &-  & 71.3 & -& 65.4 & - & 75.1 & - & 73.9 &- \\ \cdashline{2-16}[0.5pt/5pt]
        ~ & BadNets & 71.7 & 99.9 & 66.2 & 99.9 & 66.9 & 0.2 & 67.1 & 96.8 & 69.9 & 1.0 & 75.9 & 0.3 &  74.5 & 0.1 & 73.5 & 0.4 \\ 
        ~ & Blend & 72.1 & 100 & 69.4 & 0.0 & 66.7 & 0.3 & 67.8 & 97.4 & 69.3 & 26.8 & 73.0 & 0.1 & 73.7 & 13.2 & 74.1 & 1.2 \\ 
        ~ & WaNet & 70.8 & 94.7 & 69.9 & 0.9 & 66.3 & 0.4 & 68.0 & 85.0 & 68.1 & 32.9 & 17.2 & 81.8 &73.9 & 0.2 & 73.3 & 0.6 \\ 
        ~ & Frequency & 76.2 & 100 & 70.6 & 0.0 & 64.1 & 100 & 70.1 & 99.3 & 70.1 & 1.4 & 75.7 & 0.1 & 75.2 & 0.5 & 74.8 & 0.0\\[0.2em]
        \rowcolor{Gray}
        \cellcolor{white} & Average & 72.7 & 98.7 & 69.0 & 25.2 & 66.0 & 25.2 & 67.6 & 93.0 & 69.4 & 15.5 & 60.5 & 20.6 & \textbf{74.3} & \textbf{3.5} & \textbf{73.9} & \textbf{0.5} \\ 
        \hline

        \multirow{6}{*}{\begin{turn}{90}ImageNet-30\end{turn}} & No Attack & 95.9 & - & 94.4 &- & 89.9 &- & 93.2 &- & 90.0 &- & 94.5 & - & 96.9 & - & 96.7 &- \\ \cdashline{2-16}[0.5pt/5pt]
        ~ & BadNets & 95.3 & 100 & 94.3 & 0.2 & 91.2 & 0.5 & 92.9 & 0.4 & 90.7 & 9.7 & 94.2 & 0.2 & 97.4 & 0.1 & 96.7 & 0.1 \\ 
        & Blend & 83.7 & 99.9 & 93.1 & 0.1 & 90.3 & 0.6 & 91.3 & 100 & 89.9 & 2.1 & 95.2 & 0.0 & 97.2 & 0.1 & 96.8 & 0.1 \\ 
        & WaNet & 93.5 & 100 & 92.0 & 1.3 & 90.5 & 0.5 & 93.8 & 99.9 & 88.8 & 2.9 & 94.5 & 0.1 & 97.3 & 0.2 & 97.1 & 0.1 \\ 
        ~ & Frequency & 92.0 & 93.3 & 92.0 & 0.3 & 88.8 & 0.4 & 91.3 & 96.5 & 87.7 & 3.9 & 94.3 & 0.4 & 96.8 & 0.1 & 96.6 & 0.1\\[0.2em]
        \rowcolor{Gray}
        \cellcolor{white} & Average & 91.1 & 98.3 & 92.9 & 0.5 & 90.2 & 0.5 & 92.3 & 74.2 & 89.3 & 5.5 & 94.6 & 0.2 & \textbf{97.2} & \textbf{0.1} & \textbf{96.8}& \textbf{0.1}\\  
        \hline
    \end{tabular}%
    \caption{
    Accuracy (ACC) and attack success rate (ASR) on three standard datasets. VIBE consistently outperforms all previous defenses across all attacks. The results are averaged over three runs and the variance does not exceed 0.2\%.}
\label{tab:main_results}
\end{table*}

\noindent
\textbf{Baseline defenses.}~We consider five state-of-the-art defenses.
Anti-backdoor learning (\textbf{ABL})~\cite{li2021neurips} 
first isolates poisoned examples and then uses them to break the correlation between the trigger and the target class.
Decoupling based defense (\textbf{DBD})~\cite{huang2022iclr} preserves the labels for samples with low training loss of a linear classifier atop self-supervised features and proceeds with semi-supervised end-to-end fine-tuning.
Causality-inspired backdoor defense (\textbf{CBD})~\cite{zhang2023cvpr} trains a poisoned model to capture the confounding effects of triggers and corrects them in subsequent classifier training.
Backdoor defense via adaptive splitting (\textbf{ASD})~\cite{gao2023cvpr} dynamically partitions the dataset into clean and poisoned subsets.
The two subsets are then used for semi-supervised training.
Victim and Beneficiary
(\textbf{VaB}) \cite{zhu23iccv} trains a victim model on a poisoned data subset.
The victim model is then utilized for semi-supervised training of the clean model. 
More details are in Appendix~\ref{app:defenses_details}.
In the transfer learning setup, we also consider the \textbf{K-Means} clustering atop frozen features as an unsupervised baseline,  \textbf{logistic regression} as a supervised counterpart, and
a \textbf{zero-shot} CLIP-style baseline that relies on encoded textual class descriptions~\cite{radford21icml}.

\noindent
\textbf{Implementation details.}
We pre-train ResNet-18 \cite{he16cvpr} feature extractor with self-supervised objective All4One \cite{estepa2023iccv} on the poisoned dataset of interest.
We then train with VIBE objective for $30k$ iterations with the proposed EM algorithm.
In every iteration, we perform the M-step using SGD.
We perform E-step every $T=1$k iterations on a sufficiently large training subset by running a CUDA-accelerated implementation of entropy-regularized optimal transport.
Transfer learning experiments involve ViT-G/14 \cite{dosovitskiy21iclr} pretrained with DINOv2 \cite{oquab24tml}.
Other details are in Appendix~\ref{app:vibe_impl_details}. 
Our code is publicly available\footnote{https://github.com/ivansabolic/VIBE}.

\noindent\textbf{VIBE models.}
We experimentally validate two model variants.
VIBE-SelfSupervised (\textbf{VIBE-SS}) uses a randomly initialized feature extractor that we first pre-train with self-supervision \cite{estepa2023iccv} on the poisoned dataset.
Then, we append the classification heads (\ref{subsec:paramet_post}) and optimize $\ell_\text{ELBO}$.
This is our main model.
Additionally, we consider 
VIBE-FoundationModel (\textbf{VIBE-FM}) that appends classification heads atop an off-the-shelf frozen feature extractor.
In this transfer learning setup, we keep the extractor frozen and optimize the remaining parameters.
We validate both models with the full factorization $p(y|\vec{x}, l)$ and the approximation $p(y|l)$ as denoted with  (\textbf{F}) and (\textbf{A}) respectively.

\section{Experimental results}

\textbf{Resilience to backdoor attacks.} Table \ref{tab:main_results} compares VIBE-SS against five baseline defenses on three standard benchmarks.
The averaged performance over all attacks indicates that VIBE-SS outperforms all baselines by a large margin.
In particular, the absolute ASR improvement of VIBE-SS-F over the best baseline on CIFAR-10 (ABL) is more than $12$pp.
Similarly, VIBE-SS-F achieves over $14$pp ASR improvement over the best baseline ASD on CIFAR-100. 
Finally, both versions of VIBE-SS attain $0.1\%$ ASR on ImageNet-30, resulting in almost complete resilience to the considered attacks.
These improvements in robustness
do not impact clean label accuracy (ACC), which does not hold for previous methods.

Interestingly, our experiments reveal failure modes in all existing baselines.
For example, ABL and DBD are ineffective against the Adap-Blend attack, while VAB does not defend against the all-to-all attack.
Likewise, ASD fails against Frequency and Adap-Style attacks.
In contrast, VIBE-SS-F demonstrates near-complete resilience to all attacks except Adap-Patch and Adap-Blend, 
while still outperforming the best defense with over a 20pp ASR improvement for the latter.

\noindent
\textbf{Transfer learning.} 
Modular formulation of VIBE allows integration of large-scale pretrained feature extractors.
Thus, we can analyze the performance of backdoor attacks in the transfer learning setup.
We consider four relevant baselines: K-Means, logistic regression, zero-shot CLIP, and the state-of-the-art defense DBD.
DBD relies on self-supervised features, so it fits well within this setup.
Table \ref{tab:dinov2} indicates that 
VIBE-FM consistently outperforms the considered baselines across all attacks.
In particular, VIBE-FM-A delivers a complete ASR resilience on CIFAR-100 and only $0.1\%$ ASR on CIFAR-10, while VIBE-FM-F attains only slightly worse results.
Again, improved resilience comes without impact on the clean label accuracy.
Thus, VIBE framework is effective even with foundation models.
Interestingly, K-Means and the zero-shot baseline exhibit considerable resilience due to not training with corrupted labels.
Still, both of them underperform in terms of accuracy,
which emphasizes the importance of labels even for powerful feature extractors.
Logistic regression is more vulnerable than K-means 
due to naive training on corrupted labels.
Evaluating backdoor defenses in combination with frozen backbones is becoming increasingly important with the advent of robustly trained foundation models \cite{yang24icml, xun2024arxiv}.
\begin{table}[htb]
\setlength{\tabcolsep}{1.5pt}
    \footnotesize
    \centering
\begin{tabular}{cccccccc|cccc} 
    \hline
    \multirow{2}{*}{Data} & Def $\rightarrow$  &
     \multicolumn{2}{c}{LogReg} &
     \multicolumn{2}{c}{Zero-shot} &
     \multicolumn{2}{c|}{\text{DBD}} &
     \multicolumn{2}{c}{V-FM-A} &
     \multicolumn{2}{c}{V-FM-F} \\
     & Att $\downarrow$ & ACC & ASR &  ACC & ASR & ACC & ASR & ACC & ASR & ACC & ASR \\ 
     \hline
       \multirow{7}{*}{\begin{turn}{90}CIFAR-10\end{turn}} & BNets & 97.4 & 5.2 & 94.2 & 0.4 & 99.3 & 0.1 & 99.3 & 0.0 & 99.3 & 0.0 \\ 
       & Blend & 97.3 & 17.8 & 94.2 & 0.5 & 99.3 & 0.1 & 99.3 & 0.1 & 99.3 & 0.1 \\ 
       & WaNet & 97.4 & 5.2& 94.2 & 0.5 & 99.3 & 0.1 & 99.3 & 0.1 & 99.3 & 0.1 \\ 
       & Freq & 97.6 & 2.9& 94.2 & 0.3 & 99.3 & 0.0 & 99.2 & 0.0 & 99.3 & 0.0 \\
       & Patch & 99.0 & 0.2& 94.2 & 0.2 & 99.3 & 0.1 & 99.2 & 0.1 & 99.3 & 0.0 \\ 
       & Blend & 99.0 & 15.5& 94.2 & 0.6 & 99.4 & 20.5 & 99.2 & 0.6 & 99.3 & 0.9 \\ 
       & LC & 99.1 & 0.2 & 94.2 & 0.2 & 99.3 & 0.2& 99.3 & 0.1 & 99.3 & 0.1 \\ 
        \rowcolor{Gray}
        \cellcolor{white} & Avg. & 98.1 & 6.7 & 94.2 & 0.4 & \textbf{99.3} & 3.0 & \textbf{99.3} & \textbf{0.1} & \textbf{99.3} & \textbf{0.2} \\ 

     \hline
      \multirow{4}{*}{\begin{turn}{90}CIFAR-100\end{turn}} & BNets & 63.6 & 66.6 & 74.1 & 0.3 & 90.7 & 6.7 & 92.3 & 0.0 & 91.6 & 0.1 \\ 
       & Blend & 63.0 & 66.3 & 74.1 & 0.4& 90.5 & 8.5 & 92.3 & 0.0 & 91.5 & 2.1 \\ 
       & Wanet & 57.9 & 52.4 & 74.1 & 0.5 & 90.8 & 0.1 & 92.2 & 0.0 & 91.6 & 0.1 \\ 
       & Freq& 57.9 & 45.6 & 74.1 & 0.2 & 90.7 & 0.0 & 92.2 & 0.0 & 91.6 & 0.0\\[0.2em]
        \rowcolor{Gray}
       \cellcolor{white} & Avg. &  60.6 & 57.7 & 74.1 & 0.4 & 90.7 & 3.8 & \textbf{92.3} & \textbf{0.0} & \textbf{91.6} & 0.6 \\ 
       \hline
    \end{tabular}
    \caption{VIBE performance atop large-scale pretrained models.  } 
    \label{tab:dinov2}
\end{table}

\noindent
\textbf{Large-scale evaluation.}
The standard evaluation benchmarks for backdoor attacks consider datasets with a relatively small class count.
Thus, we further consider a large-scale setup on the ImageNet-1k dataset.
We consider the standard attacks BadNets, Blend and WaNet, 
as well as a 
universal backdoor attack (UBA) \cite{schneider2024iclr} that is specifically tailored for 
targeting many classes at once.  
Table \ref{tab:dinov2_imagenet1k} shows that VIBE-FM with DINOv2 consistently outperforms relevant baselines and attains near complete resilience to the considered attacks. 
For reference, baseline defense DBD fails in the case of the Blend attack and yields almost 1.4pp lower accuracy.
Both logistic regression and K-Means attain lower accuracies and higher attack success rates.
Interestingly, the zero-shot baseline achieves competitive resilience of $0.1\%$ at the cost of poor accuracy.
This analysis shows that VIBE-FM is beneficial in large-scale setups.
\begin{table}[htb]
    \footnotesize
    \setlength{\tabcolsep}{1pt}
    \centering
\begin{tabular}{ccccccccc|cccc} 
    \hline
     Method & \multicolumn{2}{c}{K-Means} & 
     \multicolumn{2}{c}{LogReg} &
     \multicolumn{2}{c}{Zero-shot} &
     \multicolumn{2}{c|}{\text{DBD}} &
     \multicolumn{2}{c}{V-FM-A} &
     \multicolumn{2}{c}{V-FM-F} \\
     Attack & ACC & ASR &  ACC & ASR & ACC & ASR & ACC & ASR & ACC & ASR & ACC & ASR  \\
     \hline
        BNets & 65.0 & 1.6 & 78.1 & 4.0 & 69.2 & 0.0 & 80.9 & 0.1 & 82.9 & 0.0 & 81.1 & 0.0 \\ 
        Blend & 65.0 & 1.9 & 78.6 & 9.1 & 69.0 & 0.1 &  81.5 & 4.3 & 83.1 & 0.0 & 81.5 & 0.1  \\ 
        WaNet & 64.9 & 1.6 & 78.4 & 5.1 & 69.0 & 0.1 & 81.6 & 0.2 & 83.1 & 0.0 & 81.5 & 0.1  \\ 
        UBA-P & 65.3 & 3.2 & 79.5 & 0.1 & 69.0 & 0.1 & 82.0 & 0.1 & 82.8 & 0.1 & 81.3 & 0.1 \\ 
        UBA-B & 65.5 & 3.2 & 79.4 & 0.1 & 69.0 & 0.1 & 81.5 & 0.1 & 83.0 & 0.1 & 81.3 & 0.1 \\ 
        \rowcolor{Gray}
        Avg. & 65.3 & 2.3 & 78.8 & 3.7 & 69.0 & 0.1 & 81.6 & 1.0 & \textbf{83.0} & \textbf{0.0} & 81.3 & \textbf{0.1} \\ 
     \hline
    \end{tabular}
    \caption{VIBE-FM performance on the ImageNet-1k dataset.} 
    \label{tab:dinov2_imagenet1k}
\end{table}

In the case of large-scale evaluation with VIBE-SS, we use ResNet-50 feature extractor pretrained on poisoned instances of the ImageNet-1k dataset.
Again, VIBE-SS attains significantly higher accuracy than baseline DBD while keeping ASR at $0.1\%$, as detailed in Appendix \ref{app:in1k_vibe_ss}.

\begin{figure}[htb]
    \centering    \includegraphics[width=0.95\linewidth]{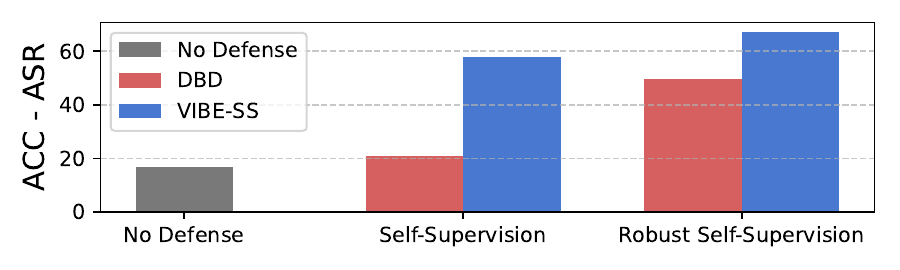}
  \caption{
  VIBE defends against the attacks on self-supervision.
  }
    \label{fig:selfsup_backdoor}
\end{figure}

\noindent
\textbf{Attacks on self-supervision.}
VIBE framework relies on feature extractor pre-training.
Thus, we analyze robustness against backdoor attacks \cite{saha22cvpr,li23iccv} that target self-supervised objective. 
Figure \ref{fig:selfsup_backdoor} compares VIBE-SS with the DBD baseline on CIFAR-10 poisoned with the CTRL attack \cite{li23iccv}.
VIBE outperforms the DBD baseline when built atop the standard SimCLR \cite{chen2020simple} pre-training and its robust counterpart MIMIC \cite{han2024mutual}, as detailed in Appendix~\ref{app:selfsup_attacks}.

We further devise our adaptive attack that targets All4One \cite{estepa2023iccv} pre-training objective used in the main experiments.
We construct a trigger that moves feature representations of 
poisoned examples towards the target classes.
VIBE successfully defends against this adaptive attack on the CIFAR-10 dataset with accuracy of $94.5\%$ 
and ASR of $0.6\%$.
Details of the attack are in Appendix~\ref{app:adaptive attacks}.


\noindent
\textbf{Combining multiple attacks.}
Existing evaluation benchmarks consider every backdoor attack in isolation.
We further harden the task by applying multiple backdoor attacks to the same instance of the CIFAR-10 dataset.
In particular, we inject visible patch attack BadNets and the clean-label attack LC.
We then evaluate robustness against each attack independently and the combined attack.
Table \ref{tab:multiple_attacks} shows that VIBE can successfully defend against combined attacks, while 
the filtering strategy of the DBD baseline fails.
\begin{table}[ht]
\footnotesize
\setlength{\tabcolsep}{2pt}
    \centering
\begin{tabular}{ccccc} 
    \hline
     Method & ASR (BadNets) & 
     ASR (LC) &
     ASR (BN \& LC) &
     ACC \\
     \hline
        DBD-SS & 99.7 & 99.8 & 99.8 & 79.1 \\
        VIBE-SS-A & 1.2 & 1.7 & 1.3 & 93.8 \\
        VIBE-SS-F & 1.8 & 2.2 & 2.0  & 93.5  \\
     \hline
    \end{tabular}
    \caption{VIBE performance on combined attacks. } 
    \label{tab:multiple_attacks}
\end{table}

\begin{figure}[ht]
    \centering
    \includegraphics[width=\linewidth]{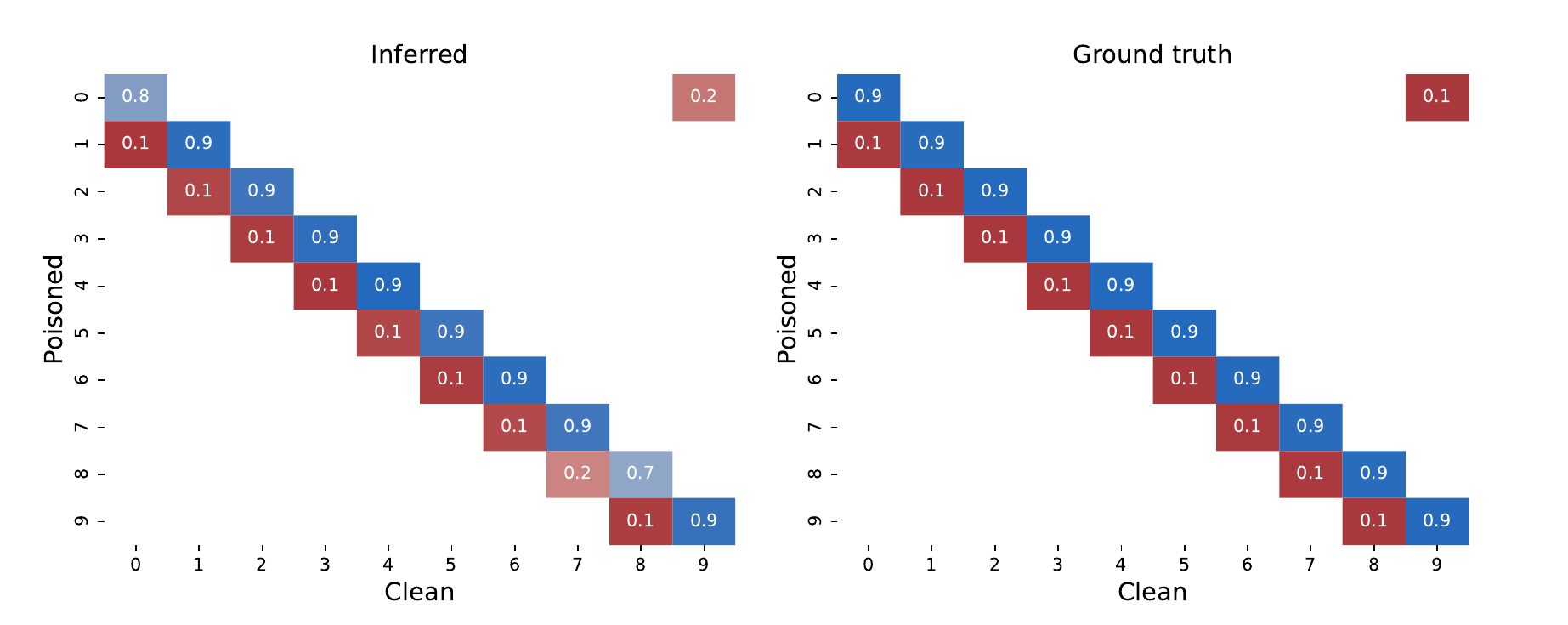}
    \caption{Inferred poisoning distributions $p(y|l)$ for all-to-all attack
    on CIFAR-10 (left) and the corresponding groundtruth (right).}
    \label{fig:p_y_given_l}
\end{figure}

\noindent
\textbf{Inferring attacker behavior.}
VIBE with approximate factorization can seamlessly recover class poisoning patterns by analyzing $p(\underline{y}|\underline{l})$ for every combination of $y$ and $l$.
To showcase this, we consider BadNets all-to-all attack that poisons all the classes in the CIFAR-10 dataset.
Figure \ref{fig:p_y_given_l} visualizes the inferred poisoning patterns (left) that resemble the actual patterns (right).
In the case of full factorization (\ref{eq:pois_post}), poisoning rules can be recovered by marginalization.
This property emerges from the VIBE formulation
and may not be easily recovered with previous defenses.

\noindent\textbf{Computational requirements.}
Our main experiments are conducted on a single NVIDIA RTX A4500 with 20GB of RAM.
Table~\ref{tab:mem} shows that VIBE training necessitates similar computational requirements as previous works, facilitating reproducibility. 
Furthermore, VIBE converges $3\times$ faster than the DBD baseline on the full ImageNet-1k dataset.
In this case, the entropy-regularized optimal transport from E-step requires 10.5GB of GPU RAM and takes 36.5 seconds. 
Still, E-step is ran only 30 times throughout the training,  keeping the runtime feasible. 
\begin{table}[ht]
\centering
\footnotesize
\setlength{\tabcolsep}{1pt}
\begin{tabular}{lccccccccccc}
\hline
Def $\rightarrow$ & \multicolumn{2}{c}{ASD \cite{gao2023cvpr}} & \multicolumn{2}{c}{VaB \cite{zhu23iccv}} & \multicolumn{2}{c}{self-sup} & \multicolumn{2}{c}{+DBD \cite{huang2022iclr}} & \multicolumn{2}{c}{+VIBE-SS} \\
Data $\downarrow$ & Mem & Time  & Mem & Time  & Mem & Time & Mem  & Time & Mem  & Time \\ \hline
C-100 & 4.0GB & 3.8h & 1.8GB & 2.7h & 2.1GB  & 8.5h & 2.7GB & 5.4h  & 1.6GB & 0.9h \\
IN-30 & 3.7GB & 7.2h & 7.6GB & 51.5h & 4.1GB & 7.0h & 4.8GB & 7.7h & 5.9GB & 2.3h \\
\hline
\end{tabular}
\caption{Computational requirements of VIBE-SS.}
\label{tab:mem}
\end{table}

\section{Discussion}
\begin{figure}[htb]
    \centering
    \includegraphics[width=\linewidth]{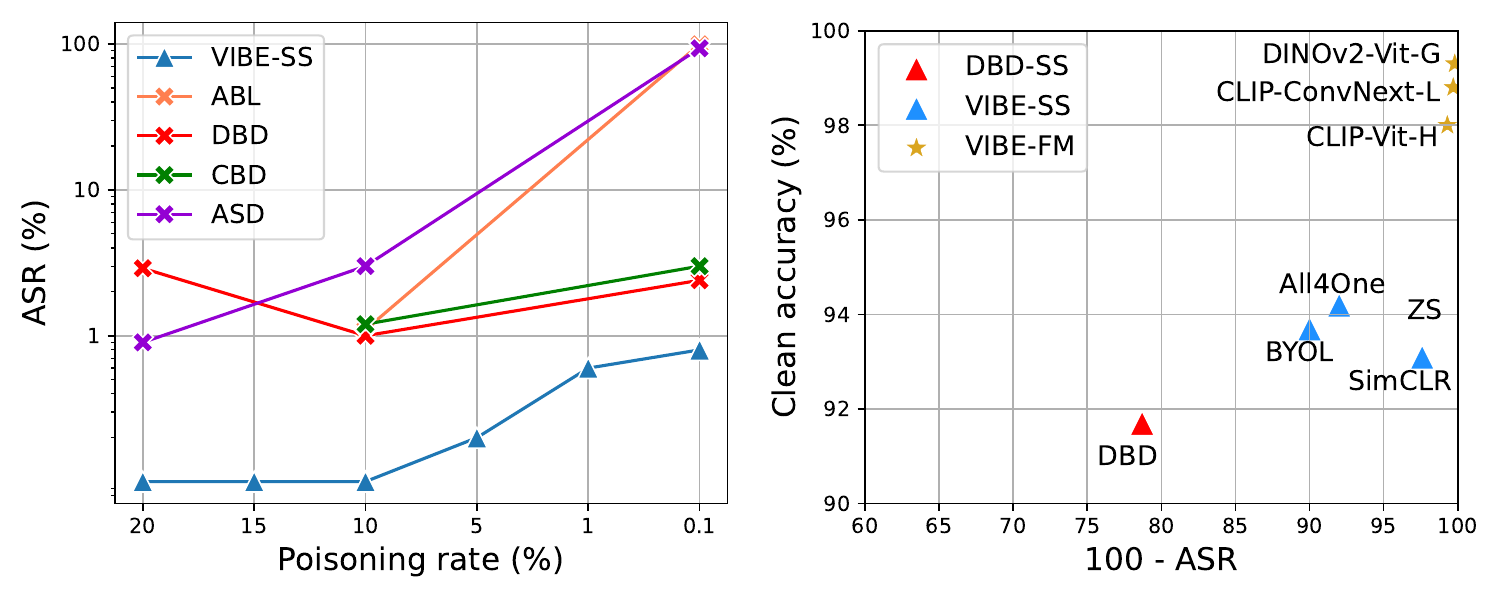}
    \caption{
    VIBE preserves strong performance for different poisoning rates (left) and can be pre-trained with different self-supervised objectives as well as built atop foundation models (right). 
    }
    \label{fig:ablation_poisoning_selfsup}
\end{figure}

\textbf{On different poisoning rates.}
Backdoor attacks typically drop the poisoning rate $\gamma$ in order to obstruct the defense.
Decreasing the poisoning rate $\gamma$ does not affect
VIBE since it simplifies the posterior recovery 
due to better overall alignment 
of the observed $\underline{y}$ and the latent $\underline{l}$.
The left side of 
Figure \ref{fig:ablation_poisoning_selfsup} shows the attack success rate in log-scale for different poisoning rates on BadNets-poisoned CIFAR-10.
While other baselines lose their performance with low poisoning rate, VIBE remains robust. 
The strong performance across different poisoning rates can be attributed to accurate pseudolabels. 
Our pseudolabels match 99\% of clean labels on CIFAR-10 and 95\% of clean labels on CIFAR-100.

\noindent
\textbf{On the choice of feature extractor.}
VIBE can be built atop different self-supervised pre-training objectives and frozen feature extractors.
The right side of
Figure \ref{fig:ablation_poisoning_selfsup} shows the average performance over six attacks on CIFAR-10.
VIBE-SS and VIBE-FM deliver competitive results in all cases.

\noindent
\textbf{On the impact of preprocessing.}
The proposed preprocessing strategy (\ref{sec:clean_label_manifold}) removes samples outside the training manifold.
This design decision improves performance on clean label attacks without hampering generalization or the performance on other attack types.
For example, data preprocessing reduces the LC attack ASR from $14\%$ to $6.3\%$ on the CIFAR-10 dataset.
Similar performance gains can be observed for other clean-label attacks \cite{yu2024icml,barni2019ieee}, as shown in Appendix \ref{app:result_clean_label}.
Furthermore, improvement in resilience does not affect clean label accuracy significantly.



\noindent
\textbf{On hyper-parameter sensitivity.} 
We validate VIBE performance for different temperatures $\nu$, $\kappa$ and $c$, E-step frequency $T$, values of the learning rate, distance thresholds $\delta$ and entropy regularization $\lambda$ in Appendix~\ref{app:hyper_sensitivity}.
VIBE performance is consistent across different hyper-parameter values.
 
\section{Conclusion}
We have presented VIBE, the first backdoor defense that views clean labels as unobserved latent variables.
We frame the training of a clean classifier as a latent posterior recovery problem and show how to efficiently solve it through expectation maximization (EM).
Specifically, our E-step infers clean pseudolabels by solving an entropy-regularized optimal transport problem via the computationally efficient matrix scaling algorithm \cite{cuturi13neurips}.
Our M-step conducts gradient descent updates on the model parameters that are pre-trained with self-supervised objective on the poisoned dataset to improve convergence \cite{huang2022iclr}.
Our experiments indicate that VIBE-SS
provides substantial defense against all considered 
backdoor attacks and remains effective against both adaptive and combined attacks.
Being modular, VIBE can also incorporate off-the-shelf foundation models and attain strong performance in
this increasingly relevant setup.

\newpage
\noindent
\textbf{Acknowledgments.} This work has been co-funded by the European Defence Fund grant
EICACS and Croatian
Recovery and Resilience Fund - NextGenerationEU (grant C1.4 R5-I2.01.0001). 


{
    \small
    \bibliographystyle{ieeenat_fullname}
    \bibliography{main}
}

\clearpage
\setcounter{page}{1}
\appendix
\onecolumn
\begin{center}
  {\Large \textbf{Seal Your Backdoor with Variational Defense}} \\[1em]
    {\large Supplementary Material}
\end{center}


\section{Detailed derivation of VIBE E-step}
\label{app:derivation_estep}
Given a set of trainable parameters $\theta$, our expectation step infers outputs of approximate clean class posterior $q$ for dataset examples as:
\begin{align}
   \frac{1}{N} \ell_\text{ELBO}(q|\theta, \mathcal{D}) &= \frac{1}{N} \sum_{i=1}^N \mathbb{E}_{l^i \sim q(\cdot|\vec{x}^i, y^i)} \left[ \ln p_{\theta}(y^i | l^i, \vec{x}^i) + \ln p_\theta(l^i | \vec{x}^i) - \ln q(l^i|\vec{x}^i, y^i) \right] \\
   &= \sum_{i=1}^N \sum_{l=1}^K \underbrace{\frac{1}{N} q(l|\vec{x}^i, y^i)}_{\mathbf{Q}_{i, l}} \ln [\underbrace{p_\theta(y^i | l, \vec{x}^i) p_\theta(l | \vec{x}^i)}_{\mathbf{P}_{i, l}}] - \frac{1}{N} q(l|\vec{x}^i, y^i)\ln \left[ N   \frac{1}{N} q(l|\vec{x}^i, y^i) \right] \\
    &= \sum_{i=1}^N \sum_{l=1}^K \mathbf{Q}_{i, l} \ln \mathbf{P}_{i, l}  - \sum_{i=1}^N \sum_{l=1}^K\mathbf{Q}_{i, l} \ln \mathbf{Q}_{i, l} - \sum_{i=1}^N \sum_{l=1}^K \mathbf{Q}_{i, l} - \ln N \\
    &= \text{tr} (\mathbf{Q}^\top \ln \mathbf{P})  + \mathbb{H}(\mathbf{Q}) + 1 - \ln N \\
    &\geq \text{tr} (\mathbf{Q}^\top \ln \mathbf{P})  + \frac{1}{\lambda} \mathbb{H}(\mathbf{Q}) + \underbrace{1 - \ln N }_{\text{const.}} \label{eq:otem}
\end{align}
Here, we defined the discrete entropy of a transport matrix $\mathbb{H}(\mathbf{Q})$  as $\mathbb{H}(\mathbf{Q}) := - \sum_{i,j} \mathbf{Q}_{i,j} (\ln \mathbf{Q}_{i,j} - 1) = \text{tr}(\mathbf{Q}^\top (1- \ln \mathbf{Q}))$, following \cite{peyre19ftml}.
The inequality holds since $\lambda > 1$ while $ 1 - \ln N$ is a constant dependent on training dataset size.

We observe that by the definition of $\mathbf{Q}$ each matrix row $\mathbf{Q}_{i,:}$ sums to $1/N$. 
Further assuming some prior over classes $\boldsymbol{\pi}$, the set of all possible solutions $\mathbf{Q}$ of the objective (\ref{eq:ot}) forms a polytope:
\begin{equation}
    \mathcal{Q}[\boldsymbol{\pi}] = \{ \, \textbf{Q} \in \mathbb{R}^{N \times K}_+ \, | \, \textbf{Q}^\top \mathbf{1}_N = \boldsymbol{\pi}, \,\,  \textbf{Q} \mathbf{1}_K = \frac{1}{N} \mathbf{1}_N \, \}.
\end{equation}

Solving the objective (\ref{eq:otem}) on polytope $\mathcal{Q}[\boldsymbol{\pi}]$ corresponds to an entropy-regularized optimal transport problem:
\begin{align}
    \underset{\mathbf{Q} \, \in \, \mathcal{Q}[\boldsymbol{\pi}]}{\max} \,\,\, \text{tr} (\mathbf{Q}^\top \ln \mathbf{P})  + \frac{1}{\lambda} \mathbb{H}(\mathbf{Q}) -  \ln N + 1 = \underset{\mathbf{Q} \, \in \, \mathcal{Q}[\boldsymbol{\pi}]}{\min} \,\,\, - \text{tr} (\mathbf{Q}^\top \ln \mathbf{P})  - \frac{1}{\lambda} \mathbb{H}(\mathbf{Q})
\end{align}
Solving entropy-regularized optimal transport problems can be conveniently done by the Sinkhorn-Knopp matrix scaling algorithm \cite{cuturi13neurips,peyre19ftml}.
For completeness, we next revisit the matrix scaling algorithm.

\section{Sinkhorn-Knopp matrix scaling algorithm for VIBE objective}
\label{app:sinkhorn}
We revisit the solution to the entropy-regularized optimal transport problem via Sinkhorn-Knopp matrix scaling algorithm \cite{peyre19ftml} in the case of cost matrix  $\mathbf{M} = -\ln \mathbf{P}$ and polytope $\mathcal{Q}[\boldsymbol{\pi}]$.
Let $\vec{a} \in \mathbb{R}^N$ and $\vec{b} \in \mathbb{R}^K$ be two dual variables that correspond to the polytope constraints  (\ref{eq:polytope}).
We can write the corresponding Lagrangian as:
\begin{equation}
    L(\mathbf{Q}, \vec{a}, \vec{b}) = - \text{tr} (\mathbf{Q}^\top \ln \mathbf{P})  - \frac{1}{\lambda} \mathbb{H}(\mathbf{Q}) + \vec{a}^\top(\textbf{Q} \mathbf{1}_K - \frac{1}{N} \mathbf{1}_N) + \vec{b}^\top(\textbf{Q}^\top \mathbf{1}_N - \boldsymbol{\pi}).
\end{equation}
Rewriting $L(\mathbf{Q}, \vec{a}, \vec{b})$ using matrix trace operators and following trace rules gives us:
\begin{equation}
    L(\mathbf{Q}, \vec{a}, \vec{b}) = - \text{tr} [(\ln \mathbf{P})^\top \mathbf{Q}]  + \frac{1}{\lambda} \text{tr}[(\ln \mathbf{Q} - 1)^\top \mathbf{Q}] + \text{tr}[(\vec{a}\mathbf{1}_K^\top)^\top\textbf{Q}] - \text{tr}[\vec{a}^\top \frac{1}{N} \mathbf{1}_N] + \text{tr}[  (\mathbf{1}_N\vec{b}^\top)^\top\textbf{Q}] - \text{tr}[\vec{b}\boldsymbol{\pi}^\top].
\end{equation}
Aggregating all elements with $\mathbf{Q}$ gives us:
\begin{equation}
    L(\mathbf{Q}, \vec{a}, \vec{b}) =\text{tr} \left[\left(-\ln \mathbf{P} + \frac{1}{\lambda} (\ln \mathbf{Q} - 1) +  \vec{a}\mathbf{1}_K^\top 
 + \mathbf{1}_N\vec{b}^\top\right)^\top\textbf{Q}\right] - \text{tr}[\vec{a}^\top \frac{1}{N} \mathbf{1}_N)]   - \text{tr}[\vec{b}\boldsymbol{\pi}^\top].
\end{equation}
Fixing $\mathbf{a}$ and $\mathbf{b}$ and expressing differential $dL$ in terms of $d \mathbf{Q}$ equals:
\begin{equation}
    dL = \text{tr} \left[\left(-\ln \mathbf{P} + \frac{1}{\lambda} \ln \mathbf{Q} +  \vec{a}\mathbf{1}_K^\top 
 + \mathbf{1}_N\vec{b}^\top\right)^\top d\textbf{Q}\right].
 \label{eq:dL}
\end{equation}
Recall that $dL = \langle\frac{\partial L}{\partial \mathbf{Q}}, d\mathbf{Q} \rangle = \text{tr}\left[\left(\frac{\partial L}{\partial \mathbf{Q}}\right)^\top d\mathbf{Q} \right]$, so we have precomputed the $\frac{\partial L}{\partial \mathbf{Q}}$ in  (\ref{eq:dL}).
By further setting $\frac{\partial L}{\partial \mathbf{Q}} = 0$ we get:
\begin{equation}
    \mathbf{Q} = \exp(-\lambda \cdot \vec{a}\mathbf{1}_K^\top) \odot \mathbf{P}^{\lambda} \odot \exp(-\lambda \cdot\mathbf{1}_N\vec{b}^\top) = \text{diag}(\vec{u}) \mathbf{P}^{\lambda} \text{diag}(\vec{v}). 
    \label{eq:q_ms}
\end{equation}
Here, $\vec{u} = \exp (-\lambda \cdot \vec{a})$ and $\vec{v} = \exp (-\lambda \cdot \vec{b})$ are two vectors with strictly positive elements and $\odot$ is elementwise product.
Since every solution $\mathbf{Q} \in \mathcal{Q}[\boldsymbol{\pi}]$, the following equalities hold:
\begin{equation}
    \text{diag}(\vec{u}) \mathbf{P}^{\lambda} \text{diag}(\vec{v}) \mathbf{1}_K = \frac{1}{N} \mathbf{1}_N \quad \text{and} \quad  \text{diag}(\vec{v}) (\mathbf{P}^{\lambda})^\top \text{diag}(\vec{u}) \mathbf{1}_N = \boldsymbol{\pi}.
\end{equation}
The two equalities can be further rewritten as:
\begin{equation}
    \vec{u} \odot (\mathbf{P}^{\lambda}\vec{v}) = \frac{1}{N} \mathbf{1}_N \quad \text{and} \quad  \vec{v} \odot ((\mathbf{P}^{\lambda})^\top \vec{u}) = \boldsymbol{\pi}.
\end{equation}
By finding $\vec{u}$ and $\vec{v}$ that satisfy the above conditions we will effectively recover the solution $\mathbf{Q}$ \cite{nemirovski99laa}, as indicated by the equation (\ref{eq:q_ms}).
Thus, we resort to iterative updates of the two vectors, following Sinkhorn's algorithm:
\begin{equation}
    \vec{u}_{t+1} := \frac{\mathbf{1}_N}{N \cdot \mathbf{P}^{\lambda}\vec{v}_t } \quad \text{and} \quad \vec{v}_{t+1} := \frac{\boldsymbol{\pi}}{(\mathbf{P}^{\lambda})^\top \vec{u}_t}.
\end{equation}
Running the matrix scaling algorithm on modern GPU hardware makes our approach feasible even for large-scale datasets.

\section{Detailed derivation of VIBE M-step}
\label{app:derivation_mstep}
Given approximate clean label posterior for dataset examples, we turn to optimization of parameters $\theta$.
We rewrite the maximization of objective $\ell_\text{ELBO}$ as:
\begin{align}
  \max_{\theta} \,\,\, \ell_\text{ELBO}(\theta|q,\mathcal{D}) =& \min_{\theta} \,\,\, - \sum_{i=1}^N \mathbb{E}_{l^i \sim q(\cdot|\vec{x}^i, y^i)} \left[ \ln p_{\theta}(y^i | l^i, \vec{x}^i) + \ln p_\theta(l^i | \vec{x}^i) - \ln q(l^i|\vec{x}^i, y^i) \right] \\
  =& \min_{\theta} \,\,\, \sum_{i=1}^N \sum_{l=1}^K  q(l|\vec{x}^i, y^i) \left[ -\ln p_{\theta}(y^i | l, \vec{x}^i) - \ln p_\theta(l | \vec{x}^i) + \ln q(l|\vec{x}^i, y^i) \right] \\
  =& \min_{\theta} \,\,\, \sum_{i=1}^N \sum_{l=1}^K   -q(l|\vec{x}^i, y^i) \ln p_\theta(l | \vec{x}^i) - \sum_{l=1}^K q(l|\vec{x}^i, y^i) \ln p_{\theta}(y^i | l, \vec{x}^i) - \mathbb{H}[q]\\
  =& \min_{\theta} \,\,\,  \sum_{i=1}^N \text{CE}[q(l|\vec{x}^i, y^i) \, || \, p_\theta(l|\vec{x}^i)] - \mathbb{E}_{l^i \sim q(\cdot|\vec{x}^i, y^i)} \left[ \ln p_{\theta}(y^i | l^i, \vec{x}^i) \right] - \mathbb{H}[q] \\
  \cong& \min_{\theta} \,\,\,  \sum_{i=1}^N \text{CE}[q(l|\vec{x}^i, y^i) \, || \, p_\theta(l|\vec{x}^i)] - \mathbb{E}_{l^i \sim q(\cdot|\vec{x}^i, y^i)} \left[ \ln p_{\theta}(y^i | l^i, \vec{x}^i) \right]
\end{align}
Here, $\mathbb{H}$ is the entropy term independent of the parameters and thus can be ignored.

\section{VIBE training algorithm}
\label{app:train_alg}
Algorithm \ref{alg:em} shows the pseudocode for VIBE training.
Prior to running the algorithm, we apply a preprocessing step that removes off-manifold training data, as described in Section \ref{sec:clean_label_manifold}.
\begin{algorithm}
\caption{EM algorithm for training VIBE}\label{alg:em}
\begin{algorithmic}[1]
\Require Dataset $\mathcal{D}$, E-step period T, total number of iterations $N_\text{iters}$, learning rate $\epsilon$, prior temperature $c$, entropy regularization coefficient $\lambda$
\Ensure Backdoor resilient classifier
\State $\theta \gets \text{initialization}(\mathcal{D})$ \Comment{Initilization of trainable parameters.}
\State $\mathcal{D} \gets$ remove off-manifold data from the original dataset $\mathcal{D}$ \Comment{Data preprocessing as described in Alg.~\ref{alg:preprocess}}
\State  $\mathbf{Q}$ $\gets$  SK-algorithm($- \ln \mathbf{P}, \boldsymbol{\pi}, \lambda$) that solves OT problem (\ref{eq:E_step}) 
\For{$j \,\, \text{in} \,\, \{0, \dots, N_\text{iters}\} $}
\If{$j\mod T = 0$} \Comment{Perform E-step every T iterations.}
    \State $\mathbf{P} \gets$ compute probabilities for $\mathcal{D}$ with parameters $\theta$
    \State $\mathbf{Q}$ $\gets$  SK-algorithm($- \ln \mathbf{P}, \boldsymbol{\pi}, \lambda$) that solves OT problem (\ref{eq:E_step}) 
\EndIf
\State $\mathbf{X}, \vec{y} \gets$ sample minibatch from $\mathcal{D}$ \Comment{Perform M-step.}
\State $\mathcal{L} \gets$ evaluate (\ref{eq:m_step}) for $(\mathbf{X}, \vec{y})$ and $\mathbf{Q}$
\State $\theta \gets \text{SGD}(\mathcal{L}, \epsilon)$
\EndFor
\end{algorithmic}
\end{algorithm}

\section{Parameterizing VIBE posteriors}
\label{app:derivation_posterior}

\subsection{Clean class posterior}
\label{app:clean_posteriror}
We define likelihood of the encoded input $\vec{v}^i = g_\theta(\vec{x}^i)$ conditioned on the clean class $l^i$ as  a von Mises-Fisher distribution~\cite{banerjee05jmlr}:
\begin{equation}
    p_{\theta}(\vec{v}^i | l^i)  := C_d(\kappa) \exp(\kappa \boldsymbol{\mu}_{l^i}^\top \vec{v}^i ).
\end{equation}
The vector $\boldsymbol{\mu}_{l^i} \in S^{d-1}$ sets the mean direction, the hyper-parameter $\kappa$ controls the distribution spread, while $C_d(\kappa)$ is a normalization constant \cite{banerjee05jmlr}.
We set $p_\theta(\vec{v}^i | \vec{x}^i) := \delta(\vec{v}^i - g_{\theta_\text{E}}(\vec{x}^i))$ since feature extractor $g_{\theta_\text{E}}$ encodes $\vec{x}^i$ exactly into $\vec{v}^i$.
By assuming a prior over clean classes $p_\pi(l^i)$, the clean class posterior can now be recovered with the Bayes rule:
\begin{align}
\label{eq:true_post_der}
    p_\theta(l^i | \vec{x}^i) &= \int p_{\theta}(l^i | \vec{v}^i) \, p_{\theta}(\vec{v}^i | \vec{x}^i) \,\,  d\vec{v}^i = \int p_{\theta}(l^i | \vec{v}^i) \, \delta(\vec{v}^i - g_{\theta_\text{E}}(\vec{x}^i)) \,\,  d\vec{v}^i \\
    &= p_{\theta}(l^i | \vec{v}^i = g_{\theta_\text{E}}(\vec{x}^i))\\ 
    &= \frac{ p_{\theta}(\vec{v}^i | l^i) p_\pi(l^i)}{\sum_{l'} p_{\theta}(\vec{v}^i | l') p_\pi(l')} = \frac{ p_{\theta}(\vec{v}^i | l^i) \pi_{l^i}}{\sum_{l'} p_{\theta}(\vec{v}^i | l') \pi_{l'}} = \frac{\exp( \kappa\,{\vec{v}^i}^\top \boldsymbol{\mu}_{l^i} + \ln \pi_{l^i})}{ \sum_{l'} \exp( \kappa\,{\vec{v}^i}^\top \boldsymbol{\mu}_{l'} + \ln \pi_{l'})}.
\end{align}
The mixing coefficients $\boldsymbol{\pi}$ are induced by a learnable prior over clean classes, \textit{i.e.}~a categorical distribution $p_\pi(l^i) := \pi_{l^i}$.
In practice, we set $\boldsymbol{\pi} = \sigma(c \cdot \theta_\pi)$,
where $\sigma$ is softmax activation, $c$ is a hyperparmeter,
and  $\theta_\pi \in \mathbb{R}^d$ are learnable parameters.
This reparametrization ensures both gradient updates towards $\theta_\pi$ due to closed form derivation of $d \boldsymbol{\pi} / d \theta_\pi$ as well as $\sum_{l=1}^K \pi_l = 1$.

\subsection{Corrupted class posterior}
\label{app:corrupted_posteriror}

We define the corrupted class posterior $p_\theta(y^i|l^i, \vec{x}^i)$ as  cosine similarity between the corrupted class prototypes $\theta_y = \{\boldsymbol{\eta}_1, \dots, \boldsymbol{\eta}_K \}$ and output of function $h$ that process the encoded input $\vec{v}^i$ and the clean label prototype $\boldsymbol\mu_{l^i}$: 
\begin{equation}
    p_{\theta}(y^i | l^i, \vec{x}^i) := \frac{\exp(\nu \cdot {\boldsymbol{\eta}_{y^i}}^\top h(\boldsymbol{\mu}_{l^i}, \vec{v}^i))}{ \sum_{y'} \exp(\nu \cdot {\boldsymbol{\eta}_{y'}}^\top h(\boldsymbol{\mu}_{l^i}, \vec{v}^i))}.
\end{equation}
Here, $\nu$ is a scalar hyper-parameter, while details on $h$ are deferred to implementation details.

We can approximate $p_{\theta}(y^i | l^i, \vec{x}^i)$ by removing the dependence on inputs.
The approximate corrupted class posterior equals to:
\begin{equation}
\label{eq:app_appr_ext}
    p_{\theta}(y^i | l^i) := \frac{\exp(\nu \cdot {\boldsymbol{\eta}_{y^i}}^\top \boldsymbol{\mu}_{l^i})}{ \sum_{y'} \exp(\nu \cdot {\boldsymbol{\eta}_{y'}}^\top \boldsymbol{\mu}_{l^i})}.
\end{equation}
The approximated posterior (\ref{eq:app_appr_ext}) hardens the optimization of $\ell_\text{ELBO}$ but enables seamless recovery of data poisoning rules.

\section{Preprocessing via manifold learning}
\label{app:manifold_preprocess}

The key insight behind our approach is that examples poisoned with a clean-label attack move away from the data manifold in the feature space of a self-supervised model pretrained on the poisoned dataset instance.
This behavior arises from the necessity for clean-label attacks to introduce substantial perturbations to poisoned images in order to be effective.
Consequently, we can defend against them by capturing the data manifold and removing the furthest community.
Algorithm \ref{alg:preprocess} details our preprocessing strategy.
\begin{algorithm}
\caption{VIBE preprocessing strategy}\label{alg:preprocess}
\begin{algorithmic}[1]
\Require Dataset $\mathcal{D}$, pretrained feature extractor $g_\theta$, threshold  $\delta$, number of nearest neighbors $k$, number of classes $K$.
\Ensure Postprocessed dataset $\mathcal{D}$
\State $\mathbf{F} \gets \text{compute features for dataset} \,\, \mathcal{D}$ with $g_\theta$
\State $A_k \gets \text{construct}\,\, k\text{-NN graph}\,\,\text{from}\,\,\mathbf{F}$
\State  $\mathcal{C} \gets$ discover $K$+1 communities in $A_k$ with Leiden algorithm
\For{$i \,\, \text{in} \,\, \{0, \dots, K\} $}
\State $\vec{d}[i] \gets$ average pairwise distance between the $i$-th community and other communities $\mathcal{C}/i$
\EndFor
\State  $c \gets \argmax_i \, \vec{d}[i]$ \Comment{Select the index of most distant community.}
\If{$\vec{d}[c] > \delta$} \Comment{Check if distant community is far away.}
    \State $\mathcal{D} \gets$ remove examples belonging to $c$-th community from the dataset
\Else 
    \State Keep all examples in $\mathcal{D}$
\EndIf
\end{algorithmic}
\end{algorithm}

\begin{figure}[htb]
    \centering
    \includegraphics[width=.75\linewidth]{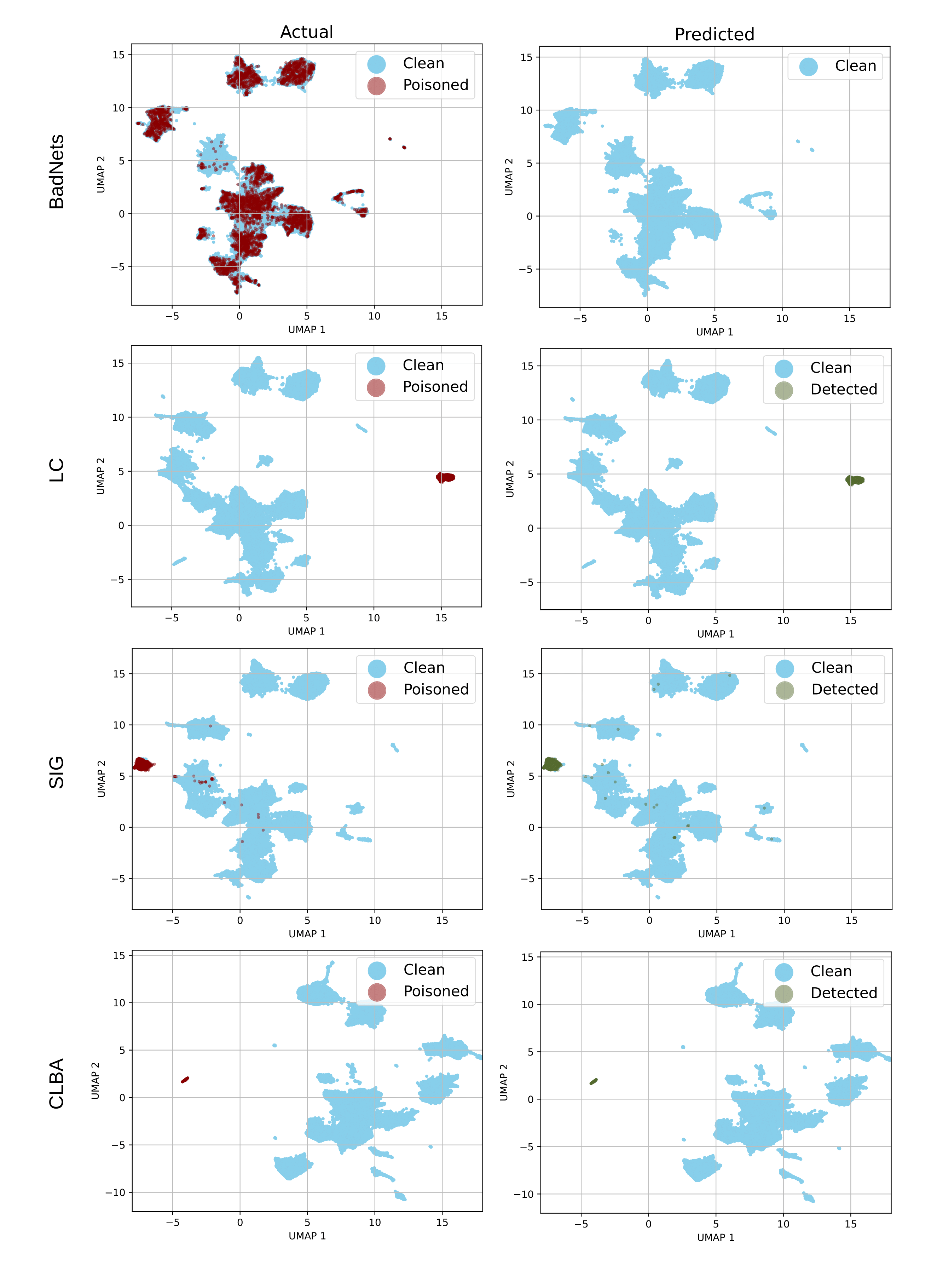}
    \caption{Preprocessing strategy successfully removes poisoned data in all considered clean-label attacks.}
    \label{fig:prepro_sig_clba}
\end{figure}

\noindent
Figure \ref{fig:prepro_sig_clba} shows the implications of the preprocessing strategy on clean-label attacks LC \cite{turner2019arxiv}, SIG \cite{barni2019ieee} and CLBA \cite{yu2024icml}, as well as poisoned-label attack BadNets \cite{gu2019ieee}.
In the presence of clean-label attacks, the captured off-manifold community corresponds to poisoned examples that are later safely removed (rows 2-4).
In the absence of clean-label attacks the captured outlier community corresponds to a fistful of outlier samples that are still close enough to the data manifold and thus are retained (row 1).
Note that the visualized distances may be distorted due to two-dimensional UMAP \cite{mcinnes2018umap-software} plots of high-dimensional space.


\section{Details on experimental setup}
This section outlines details of the considered baseline attacks and defenses.

\subsection{Details on backdoor attacks}
\label{app:attacks_details}

\textbf{BadNets.}
We follow \cite{huang2022iclr, gao2023cvpr} by adopting the same 
$2 \times 2$ trigger pattern for CIFAR-10 and CIFAR-100 and Apple logo 
for ImageNet. \\
\textbf{Blend.}
Following original work \cite{chen2017arxiv}
and \cite{huang2022iclr, gao2023cvpr}, we use 
use the \textit{Hello-kitty} trigger pattern
on CIFAR datasets and random noise pattern on ImageNet
datasets. Blending ratio is set to $0.1$. \\
\textbf{WaNet.}
We follow the setup in \cite{huang2022iclr, gao2023cvpr}
to generate trigger patterns using warping operations. 
We set $k$ to 4 on CIFAR datasets and $k$ to 224
on ImageNet datasets. \\
\textbf{Frequency.}
We rely on BackdoorBench\footnote{\url{https://github.com/SCLBD/BackdoorBench}} to reproduce Frequency
attack. All hyper-parameters are set as in \cite{wang2022eccv}. \\
\textbf{Adap-patch \& Adap-blend.}
We follow the attack setup from \cite{qi2022iclr}. \\
\textbf{Label Consistent.}
We use PGD \cite{madry2018iclr} to generate adversarial perturbations
within $L^{\infty}$ ball. Maximum magnitude $\eta$ is set to $16$, step size to $1.5$
and perturbation steps to $30$. Trigger pattern is $3 \times 3$ grid pattern in 
each corner of the image. \\
\textbf{UBA.}
We adopt the attack setup as in \cite{schneider2024iclr}.
In the Patch and Blend versions, we poison 2000 and 8000 examples 
following the original paper.

\subsection{Details on baseline defenses}
\label{app:defenses_details}
\textbf{ABL.}
To reproduce ABL, we use 
BackdoorBox\footnote{\url{https://github.com/THUYimingLi/BackdoorBox}}.
We found ABL to be very sensitive 
to its main hyper-parameter $\gamma$. Therefore, 
we conduct the grid search to find the best
$\gamma \in \left\{0, 0.1, 0.2, 0.5 \right\}$
yielding lowest ASR against every attack. We note
that this assumption might be overoptimistic in practice.
Following original work~\cite{li2021neurips}, we 
poison the model for $20$ epochs, followed by $70$ epochs of 
fine-tuning. Lastly, we unlearn the backdoor for $5$ epochs. \\
\textbf{DBD.}
We refer to the official implementation\footnote{\url{https://github.com/SCLBD/DBD}} to reproduce 
DBD and use all configurations as introduces in \cite{huang2022iclr}.
\\
\textbf{CBD.}
We use the official code implementation\footnote{\url{https://github.com/zaixizhang/CBD}} and
all configurations from \cite{zhang2023cvpr}.
\\
\textbf{ASD.}
We rely on official implementation\footnote{\url{https://github.com/KuofengGao/ASD}}
and configurations from original paper~\cite{gao2023cvpr}.
\\
\textbf{VAB.}
We rely on official code implementation\footnote{\url{https://github.com/Zixuan-Zhu/VaB}}
and follow configurations from original work~\cite{zhu23iccv}.

\section{VIBE implementation details} 

\label{app:vibe_impl_details}
We use ResNet-18 \cite{he16cvpr} feature extractor with self-supervised initialization \cite{estepa2023iccv} on the poisoned dataset of interest.
We train VIBE end-to-end for $30k$ iterations using the proposed EM algorithm.
In every iteration, we perform the M-step using SGD with learning rate $10^{-3}$ and batch size 256.
We perform a dataset-wide E-step every $T=1$k iterations by solving entropy-regularized optimal transport, with $\lambda=25$, as validated in early experiments.
Hyper-parameters $\nu$ and $\kappa$ are set to $10$ for datasets with number of classes $K \leq 30$ and to $20$ otherwise.
We fix $k=50$, $\delta=0.275$, and $c=0.02$ with
early validation experiments.
Experiments with frozen foundation model involve ViT-G/14 \cite{dosovitskiy21iclr} pretrained with DINOv2 \cite{oquab24tml}.
These experiments fix the poisoned class prototypes $\psi$ as means over poisoned labels.
Thus, we optimize clean prototypes $\phi$ for $15$k iterations using SGD with constant learning rate 
$10^{-2}$ and batch size 1024. 
We conduct E-step every $T=500$ iterations. 
In both setups, we fix $h(\boldsymbol{\mu}, \vec{v}) = \boldsymbol{\mu}+\vec{v}$, yet other choices of $h$ may also yield competitive performance. 
All experiments were conducted to maximize GPU utilization. We measured maximal memory requirements with \textit{torch.cuda.max\_memory\_allocated}.


\section{Extended results}

\subsection{Extended results for ImageNet-1k dataset}
\label{app:in1k_vibe_ss}
\begin{figure}[htb]
    \centering
    \includegraphics[width=0.5\linewidth]{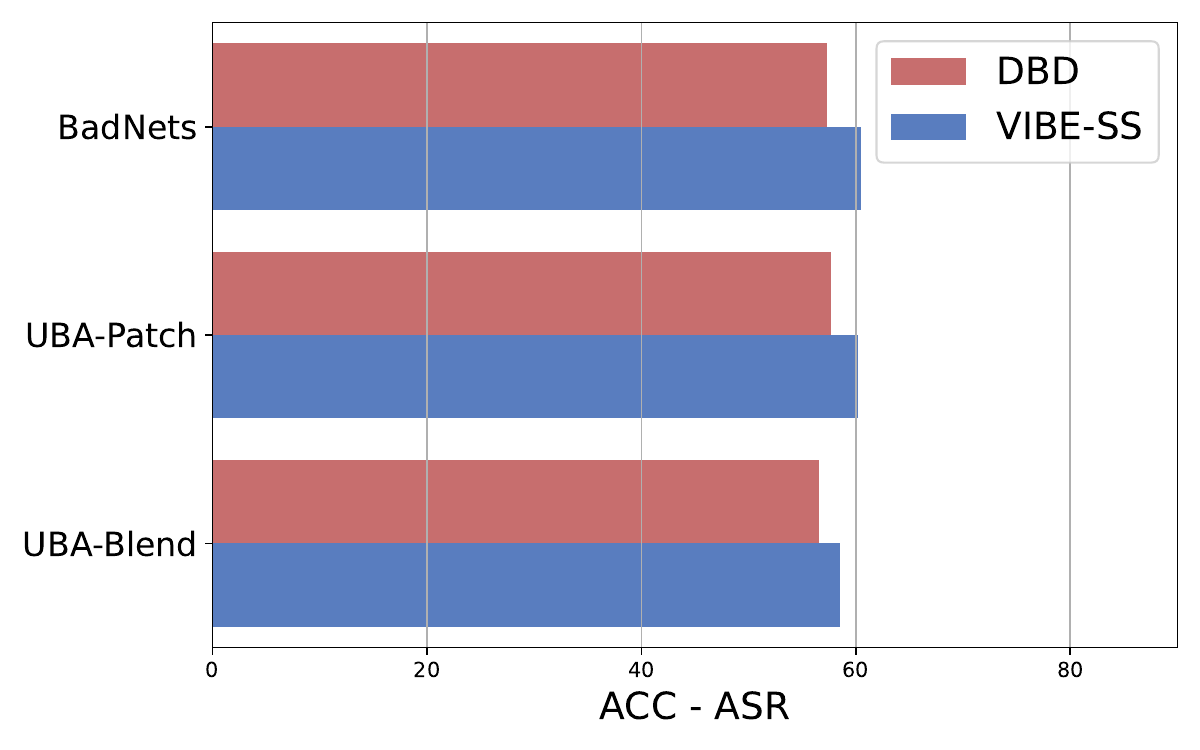}
    \caption{VIBE performance when combined with ResNet-50 pretrained on poisoned instances of the ImageNet-1k dataset.}
    \label{fig:imagenet_selfsup}
\end{figure}

\Cref{fig:imagenet_selfsup} illustrates the difference between ACC and ASR for both VIBE-SS and the DBD baseline. 
Both methods are built upon a frozen ResNet-50 pretrained on poisoned instances of the ImageNet-1k.
VIBE-SS successfully defends against a variety of attacks in this setup, achieving both high ACC and high robustness. 

\subsection{Attacks on self-supervision}
\label{app:selfsup_attacks}

Relevant backdoor attacks targeting self-supervised learning \cite{saha22cvpr, li23iccv} operate by injecting carefully designed triggers that poison the embedding space of a model pretrained according to contrastive self-supervised objectives.
In this setup, 
attack is successful if the backdoor persists even after model fine-tuning or linear probing on a small subset of clean data.
While this setup slightly differs from the one studied in our main paper,
VIBE again exhibits competitive performance.
Additionally, robust self-supervised pre-training objectives already exist \cite{bie2024mitigating, han2024mutual}. 
Modular VIBE design allows seamless integration of such robust pre-training.

Table \ref{tab:ctrl} compares VIBE performance with the most relevant baseline DBD on the CIFAR-10 dataset poisoned with CTRL \cite{li23iccv}.
In this experiment, we rely on SimCLR \cite{chen2020simple} pre-training objective to stay consistent with the original setup \cite{li23iccv}.
The first row of shows the performance of VIBE and DBD baseline against the CTRL attack \cite{li23iccv} after the standard SimCLR pretraining.
VIBE performance surpasses DBD both in terms of accuracy and ASR.
Since robust pre-training methods exist, we validate VIBE and DBD performance atop one such method MIMIC \cite{han2024mutual}.
In this case VIBE becomes even more resilient, attaining ASR of only 4\%.
In contrast, DBD suffers from a loss of generalization power on the clean data.
To understand the cause of this performance drop, we conducted a thorough hyperparameter analysis. 
We concluded that this behavior stems from the loss of labels due to the filtering strategy characteristic for DBD.
Altogether, we find robust self-supervised objectives a promising research direction that can be easily ported into the VIBE framework.

\begin{table}[ht]
\setlength{\tabcolsep}{3.5pt}
\centering
\begin{tabular}{lcc>{\columncolor[gray]{0.95}}c cc>{\columncolor[gray]{0.95}}c cc>{\columncolor[gray]{0.95}}c}
\hline
 Defense $\rightarrow$ & \multicolumn{3}{c}{LogReg} & \multicolumn{3}{c}{DBD} & \multicolumn{3}{c}{VIBE-SS} \\
Self-sup objective $\downarrow$ & ACC & ASR & ACC - ASR & ACC & ASR & ACC - ASR & ACC & ASR & ACC - ASR \\  \hline
SimCLR & 81.0 & 64.3 & 16.7 & 51.4 & 30.6 & 20.8 & 84.5 & 26.8 & \textbf{57.7} \\
SimCLR + MIMIC & 61.9 & 7.7 & 54.2 & 49.8 & 0.0 & 49.8 & 71.6 & 4.2 & \textbf{67.4} \\
\hline
\end{tabular}
\caption{Defending against the CTRL attack targeting self-supervision. Higher difference between ACC and ASR indicates better performance.}
\label{tab:ctrl}
\end{table}

\noindent
Note that we found the attacks on self-supervision to be fragile in practice. 
Consistent to the findings in \cite{li23iccv},
we were unable to reproduce the SSL attack \cite{saha22cvpr}.
Furthermore, the effectiveness of the CTRL attack \cite{li23iccv} was highly sensitive to how the poisoned data was stored.
Specifically, saving the data as \texttt{float32} preserved the attack, whereas the standard data storage format \texttt{uint8} nullified the poisoning effect.


\subsection{Adaptive attack on VIBE}
\label{app:adaptive attacks} 
Given that the attacker has knowledge about how our defense operates,
they may attempt to construct an attack which bypasses our defense mechanism. 
Such scenario is regarded as the adaptive attack. Since our defense jumpstarts
the optimization process with self-supervised initialization, one adaptive attack
would be to construct a trigger such that the self-supervised representations of 
poisoned examples resemble those of target class examples. Concretely, we optimize 
the trigger $\vec{t}$ by minimizing the distance between the representations of poisoned examples
and the target class centroid. Given the original dataset $\mathcal{D}_\text{raw}$, 
let $\mathcal{D}_\text P \in \mathcal{D} $ be the poisoned subset and 
$\mathcal{D}_\text T = \left\{ (\vec{x}, y) : (\vec{x}, y) \in \mathcal{D}, y = y_T \right\}$ 
be the target class subset. 

\begin{equation}
     \argmin_{\vec t} \frac{1}{|\mathcal{D}_\text P|}\sum_i^{|\mathcal{D}_\text{P}|} \left|\left| \mathcal{B}(g(\vec x_i), \vec t) - \frac{1}{|\mathcal{D}_\text{T}|} \sum_j^{|\mathcal{D}_\text{T}|} g(\vec{x}_j)   \right|\right|_2
\end{equation}
\begin{equation}
    \text{s.t.} \quad \left|\left| \vec{t} \right|\right|_{\infty} \leq \eps
\end{equation}
\\
\textbf{Settings.}
We conduct the attack on the CIFAR-10 dataset and set the trigger size to be the same
of the original image. $\mathcal{B}(\cdot, \vec{x})$ is a blending function with
factor $0.5$. 
We optimize for 100 iterations with SGD and base learning rate 
set to $0.1$. Learning rate is decayed with $0.1$ every $40$ iterations and  
$\eps$ is set to $\frac{32}{255}$ to conceal the trigger. Poisoned rate is $10\%$.
\\
\textbf{Results.}
For model trained with standard cross-entropy loss, the proposed adaptive attack results in ASR of $99.2\%$ and clean accuracy 
of $91.2\%$. 
On contrary, VIBE is completely robust against such attack.
It has no backdoor injected (ASR=$0.8\%$) and generalizes even
better on the clean data (ACC=$94.6\%$). 
A more comprehensive analysis indicates that VIBE corrects the labels
of poisoned examples into their actual classes.

\subsection{Extended results on clean-label attacks}
\label{app:result_clean_label}
Table \ref{tab:abl_preprocessing} empirically shows that the proposed preprocessing strategy improves VIBE performance for the most relevant clean-label attacks. 
In the case of the LC attack, VIBE offers significant robustness even without the preprocessing step.
Furthermore, the proposed preprocessing does not affect the
overall dataset size when the clean-label attacks are absent.
Thus, we attain similar performance on the poisoned-label BadNets attack.
\begin{table}[htb]
\setlength{\tabcolsep}{2pt}
    \centering
    \begin{tabular}{ccccccccc}
    \toprule
    Attack $\rightarrow$ & \multicolumn{2}{c}{BadNets} & 
     \multicolumn{2}{c}{LC} &
     \multicolumn{2}{c}{SIG} &
     \multicolumn{2}{c}{CLBA} \\ 
     Preprocess $\downarrow$ &ACC & ASR & ACC & ASR & ACC & ASR & ACC & ASR \\
    \midrule
     \ding{55} & 94.5 & 0.4 & 94.6 & 14.0 & 94.2 & 45.0 & 94.6 & 89.3 \\
    \ding{51} & 94.4 & 0.4 & 94.3 & 6.0 & 92.7 & 14.7 & 94.1 & 14.1 \\
    \bottomrule
    \end{tabular}    
    \caption{VIBE performance with and without our preprocessing strategy.}
    \label{tab:abl_preprocessing}
\end{table}

\noindent
Table \ref{tab:clean_label} compares the proposed method against other baseline defenses on the three clean-label attacks. 
On average, VIBE outperforms all considered baselines.
Furthermore, all baselines have some failure modes, such as the CLBA attack.
On the contrary, VIBE does not completely fail against any of the three clean-label attacks.
\begin{table*}[htb]
    \centering
    \footnotesize
    \begin{tabular}{ccccccccccccc}
    \toprule
          Defense $\rightarrow$ &
         \multicolumn{2}{c}{No Defense} &
         \multicolumn{2}{c}{ABL} &
         \multicolumn{2}{c}{DBD} &
         \multicolumn{2}{c}{CBD} &
         \multicolumn{2}{c}{ASD} &
         \multicolumn{2}{c}{VIBE-SS} \\ 
         Attack $\downarrow$ & ACC & ASR & ACC & ASR & ACC & ASR & ACC & ASR & ACC & ASR & ACC & ASR \\ \hline
        LC & 94.9 & 99.9 & 86.6 & 1.3 & 89.7 & 0.0 & 91.3 & 24.7 & 93.1 & 0.9 & 93.3 & 6.0 \\ 
         SIG & 92.3 & 98.0 & 81.6 & 0.1 & 91.6 & 100 & 92.5 & 1.1 & 93.4 & 1.0 & 92.7 & 14.7  \\ 
         CLBA & 93.2 & 80.0 & 84.6 & 93.4 & 91.5 & 87.6 & 90.4 & 37.8 & 93.2  & 98.0 & 94.1 & 14.1  \\ 
        \rowcolor{Gray}
        Average & 94.1 & 90.0 & 84.1 & 31.6  & 90.9 & 62.5  & 91.9  & 12.9  & 93.2 & 33.3 & \textbf{93.3} & \textbf{11.4}  \\ 
    \bottomrule
    \end{tabular}
    \caption{Extended results for clean-label attacks.}
    \label{tab:clean_label}
    \end{table*}

\noindent
Additionally, we test VIBE's resilience against targeted poisoning attacks \cite{shafahi18neurips, geiping2021iclr} which aim to misclassify one specific test example. 
Although the general goal behind these attacks slightly diverges from standard backdoor attacks, the poisoning setup is the same. 
The attacker aims to inject malicious behaviour into the model by poisoning the training data. 
Still, VIBE provides complete robustness against Witches Brew~\cite{geiping2021iclr}, a representative of targeted poisoning attacks, by reducing the ASR from 50\% to 0\%. 

\noindent
\subsection{Comparison with post-training defenses}
Unlike the training-time defenses such as VIBE, 
which operate solely on the poisoned
dataset, post-training defenses assume access to the poisoned model and additional clean data. 
Still, the outcome of both categories of defenses is a robust model.
VIBE outperforms NAD~\cite{li2021iclr} and ANP~\cite{wu2021neurips}, representatives of post-training defenses, as shown in the Table \ref{tab:post_training}.

\begin{table}[ht]
\setlength{\tabcolsep}{3.5pt}
\centering
\begin{tabular}{lcccccccc}
\toprule
 Defense $\rightarrow$ &\multicolumn{2}{c}{NAD} &\multicolumn{2}{c}{ANP} & \multicolumn{2}{c}{VIBE-SS} \\
Data $\downarrow$ & ACC & ASR & ACC & ASR & ACC & ASR \\ \midrule
CIFAR-10 & 82.9 & 6.1 & 90.9 & 2.1 & \textbf{94.1} & \textbf{1.7} \\
ImageNet-30 & 91.1 & 0.5 & - &- & \textbf{96.9} & \textbf{0.1} \\
\bottomrule
\end{tabular}
\caption{Experimental comparison with NAD~\cite{li2021iclr} and ANP~\cite{wu2021neurips}. The numbers are averaged over 4 attacks on CIFAR-10 and 3 attacks on ImageNet-30.}
\label{tab:post_training}
\end{table}

\subsection{Comparison with detection defenses}
Detection defenses focus on identifying poisoned examples.
However, retraining on data filtered by these methods may still result in poisoned models, as noted in \cite{pal2024iclr}.
Although not being the main focus of our work, VIBE can be used as a poisoned sample detector. The psuedo-labelling in E-step combined with our pre-processing strategy employed to detect clean-label data achieves performance on par with state-of-the-art detection methods SCP~\cite{guo2023iclr}, BSU~\cite{pal2024iclr} and PSBD~\cite{li2025cvpr}, as demonstrated in Table~\ref{tab:det-auroc}.

\begin{table}[ht]
\setlength{\tabcolsep}{3.5pt}
\centering
\begin{tabular}{ccccc}
\toprule
Attack $\downarrow$ & SCP & BSU & PSBD & VIBE-SS \\ \midrule
BadNets & .83/\textbf{1.0}/.30 & .95/\textbf{1.0}/.17 & \textbf{.99}/\textbf{1.0}/.10 & \textbf{.99}/.99/\textbf{.01} \\
Blend & .50/.13/.18 & .95/.99/.10 & .96/\textbf{1.0}/.14 & \textbf{.98}/.96/\textbf{.01} \\
WaNet & .73/.90/.28 & .93/.99/.10 & \textbf{.99}/\textbf{1.0}/.14 & \textbf{.99}/.98/\textbf{.00} \\
LC  & .93/.94/.18 & .95/.99/.14 & \textbf{1.0}/.99/.13 & .99/\textbf{1.0}/\textbf{.02} \\ 
\bottomrule
\end{tabular}
\vspace{-0.25cm}
\caption{Experimental comparison with detection defenses. All measurements are in AUROC/TPR/FPR format.}
\label{tab:det-auroc}
\end{table}

\subsection{Different architectures of the feature extractor}
Being agnostic to the backbone choice, VIBE can be built atop
different model architectures. Table \ref{tab:vgg11} showcases robust performance
when VGG-11 \cite{simonyan2015iclr} is used as backbone.
\begin{table}[ht]
\setlength{\tabcolsep}{3.5pt}
\centering
\begin{tabular}{lcccccccc}
\hline
 Defense $\rightarrow$ & \multicolumn{2}{c}{No defense} & \multicolumn{2}{c}{ASD}  & \multicolumn{2}{c}{VIBE-SS} \\
Attack $\downarrow$ & ACC & ASR & ACC & ASR & ACC & ASR \\ \hline
BadNets & 91.0 & 99.9 & 90.4 & 3.7  & \textbf{91.3} & \textbf{1.1} \\
Blend & 90.6 & 98.4 & 87.5 & 2.4 & \textbf{91.3} & \textbf{2.2} \\
\hline
\end{tabular}
\caption{VIBE-SS performance with VGG-11 as backbone on CIFAR-10 dataset.}
\label{tab:vgg11}
\end{table}

\section{Hyper-parameter sensitivity}
\label{app:hyper_sensitivity}
This section analyzes VIBE performance for different values of hyper-parameters.
All experiments are conducted on the CIFAR-10 dataset poisoned with BadNets attack.
VIBE attains similar performance for different hyper-parameter values.

\begin{table}[ht]
\centering
  \begin{minipage}[b]{0.33\linewidth}
\label{tab:abl_T}
\begin{center}
\begin{small}
\begin{sc}
\begin{tabular}{ccc}
\toprule
$T$ & ACC & ASR \\
\midrule
500 & 94.4 & 0.5  \\
\textbf{1000} & 94.4 & 0.4 \\
2000 & 94.4 & 0.5  \\
\bottomrule
\end{tabular}
  \caption{VIBE performance for different values of E-step period T.} 
\end{sc}
\end{small}
\end{center}
  \end{minipage}
  \hspace{0.05\linewidth}
  \begin{minipage}[b]{0.33\linewidth}
\label{tab:abl_lr}
\begin{center}
\begin{small}
\begin{sc}
\begin{tabular}{ccc}
\toprule
$lr$ & ACC & ASR\\
\midrule
$10^{-2}$  & 92.9 & 0.9  \\
$\mathbf{10^{-3}}$ & 94.4 & 0.4  \\
$10^{-4}$ & 91.0 & 0.6  \\
\bottomrule
\end{tabular}
  \caption{VIBE performance for different values of the learning rate.}
\end{sc}
\end{small}
\end{center}
  \end{minipage}
\end{table}

\begin{table}[ht]
\centering
  \begin{minipage}[b]{0.33\linewidth}
\label{tab:abl_delta}
\begin{center}
\begin{small}
\begin{sc}
\begin{tabular}{ccc}
\toprule
$\delta$ & ACC & ASR\\
\midrule
0.250 & 94.4 & 0.5 \\
\textbf{0.275} & 94.4 & 0.4  \\
0.300 & 94.3 & 0.5  \\
\bottomrule
\end{tabular}
  \caption{VIBE performance for different distance thresholds $\delta$.}
\end{sc}
\end{small}
\end{center}
  \end{minipage}
  \hspace{0.05\linewidth}
  \begin{minipage}[b]{0.33\linewidth}
\label{tab:abl_c}
\begin{center}
\begin{small}
\begin{sc}
\begin{tabular}{ccc}
\toprule
$c$ & ACC & ASR\\
\midrule
$0.05$  & 94.4 & 0.6  \\
$\mathbf{0.02}$ & 94.4 & 0.4  \\
$0.01$ & 94.4 & 0.5  \\
\bottomrule
\end{tabular}
  \caption{VIBE performance for different values of the temperature $c$ used for prior $\vec \pi = \sigma(c \cdot \theta_\pi)$.}
\end{sc}
\end{small}
\end{center}
  \end{minipage}
\end{table}
  
\begin{table}[ht]
\centering
  \begin{minipage}[b]{0.25\linewidth}
\label{tab:abl_kappa}
    \begin{center}
\begin{small}
\begin{sc}
\begin{tabular}{ccc}
\toprule
$\kappa$ & ACC & ASR\\
\midrule
20 & 94.5 & 0.3  \\
\textbf{10} & 94.4 & 0.4  \\
2 & 92.7 & 0.8  \\
\bottomrule
\end{tabular}
  \caption{VIBE performance for different values of temperature $\kappa$ used in $p_{\phi, \theta} (l | \vec{x})$.}
\end{sc}
\end{small}
\end{center}
  \end{minipage}
  \hspace{0.05\linewidth}
  \begin{minipage}[b]{0.25\linewidth}
\label{tab:abl_tau}
    \begin{center}
\begin{small}
\begin{sc}
\begin{tabular}{ccc}
\toprule
$\nu$ & ACC & ASR\\
\midrule
20 & 94.4 & 0.5  \\
\textbf{10} & 94.4 & 0.4  \\
2 & 93.3 & 0.6  \\

\bottomrule
\end{tabular}
  \caption{VIBE performance for different values of $\nu$ used in $p_{\phi, \psi} (y|l)$.}
\end{sc}
\end{small}
\end{center}
  \end{minipage}
  \hspace{0.05\linewidth}
  \begin{minipage}[b]{0.25\linewidth}
\label{tab:abl_lambda}
    \begin{center}
\begin{small}
\begin{sc}
\begin{tabular}{ccc}
\toprule
$\lambda$ & ACC & ASR\\
\midrule
10 & 94.4 & 0.5  \\
\textbf{25} & 94.4 & 0.4  \\
50 & 94.5 & 0.5 \\
\bottomrule
\end{tabular}
  \caption{VIBE performance for different entropy regularization hyperparameters $\lambda$.}
\end{sc}
\end{small}
\end{center}
  \end{minipage}
\end{table}

\section{Qualitative results}
\begin{figure}[htb]
    \centering
    \includegraphics[width=0.8\linewidth]{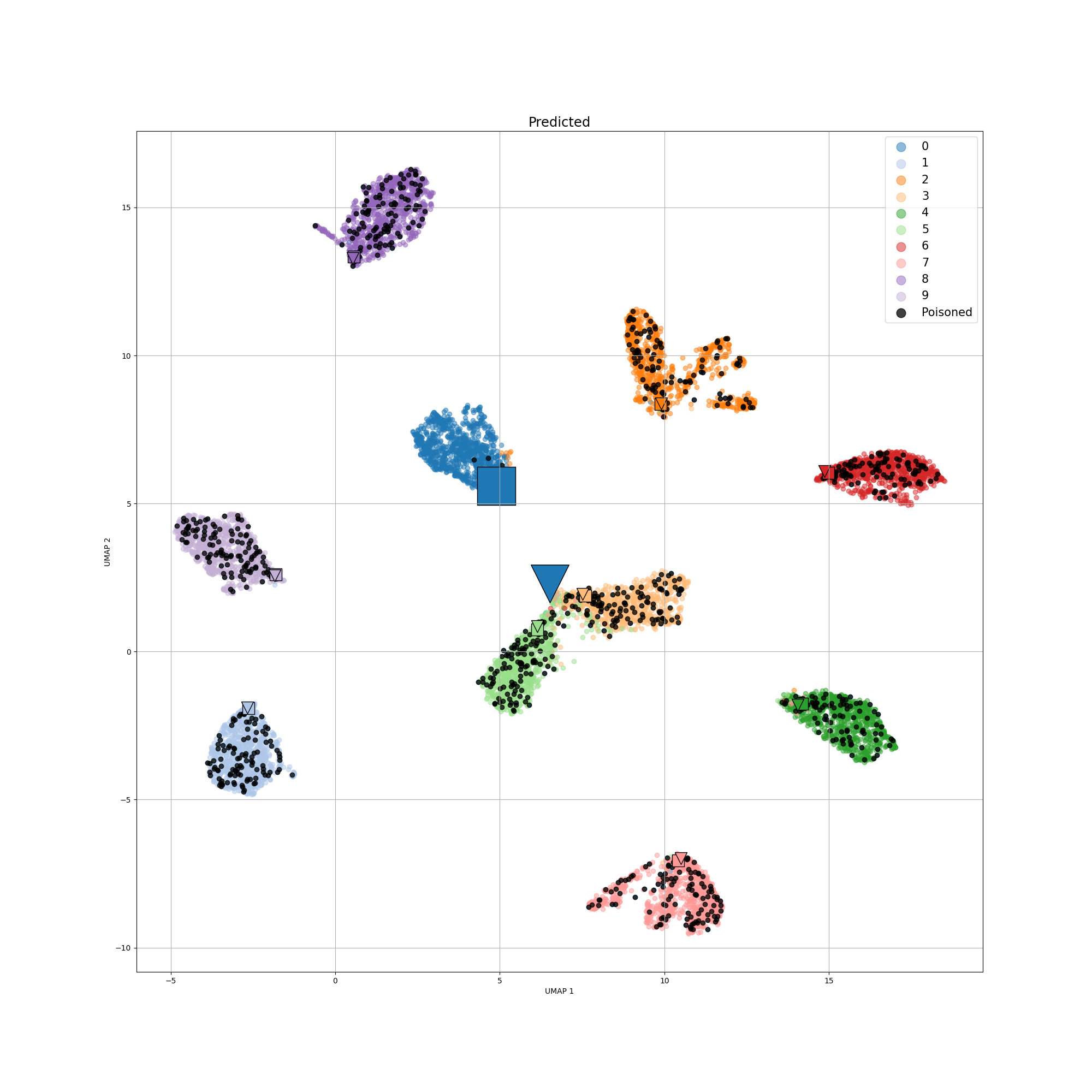}
    \caption{VIBE latent space at the end of training on \textit{all-to-one} BadNets-poisoned CIFAR-10. Target class is colored in blue. Poisoned examples having the same label as the target class are colored in black. Clean class prototypes $\mu$ are marked with squares, while poisoned class prototypes $\eta$ are marked with triangles.}
    \label{fig:latent_space}
\end{figure}

\Cref{fig:latent_space} shows the VIBE latent space after the training. 
The learned clean class prototypes are denoted with squares while the poisoned class prototypes are denoted with triangles.
Poisoned class prototype (blue triangle) of the target class is located in the center of the data manifold because the target class contains examples from all other classes in \textit{all-to-one} poisoning.
Contrary, the clean class prototypes learned by VIBE are adequately assigned across the clusters.

\end{document}